\documentclass{article}

\PassOptionsToPackage{numbers, compress}{natbib}

\usepackage[preprint]{neurips_2022}

\usepackage[utf8]{inputenc} 
\usepackage[T1]{fontenc}    
\usepackage{hyperref}       
\usepackage{url}            
\usepackage{booktabs}       
\usepackage{amsfonts}       
\usepackage{nicefrac}       
\usepackage{microtype}      
\usepackage{xcolor}         
\usepackage{scalerel}
\usepackage{tikz}
\usepackage{wrapfig}

\usepackage{graphicx}
\usepackage{doi}
\usepackage{caption}
\usepackage{subcaption}
\usepackage{amsmath}
\usepackage{amsthm}
\usepackage{verbatim}

\title{
Encoding Concepts in Graph Neural Networks
}

\author{
Lucie Charlotte Magister\thanks{Equal Contribution} \\
  University of Cambridge\\
  Cambridge, UK\\
  \texttt{lcm67@cam.ac.uk} \\
  \And
  Pietro Barbiero\footnotemark[1] \\
  University of Cambridge\\
 Cambridge, UK\\
  \texttt{pb737@cam.ac.uk} \\
  \And 
  Dmitry Kazhdan \\
  University of Cambridge\\
  Cambridge, UK\\
  \texttt{dk525@cam.ac.uk} \\
  \And
  Federico Siciliano \\
  University of Rome, La Sapienza\\
  Rome, Italy \\
  \texttt{federico.siciliano@uniroma1.it} \\
  \And
  Gabriele Ciravegna \\
  Universit\'e C\^{o}te D'Azur\\
  Le Chesnay-Rocquencourt, France\\
  \texttt{gabriele.ciravegna@inria.fr} \\
  \And
  Fabrizio Silvestri \\
  University of Rome, La Sapienza\\
  Rome, Italy\\
  \texttt{fabrizio.silvestri@uniroma1.it} \\
  \And
  Mateja Jamnik \\
  University of Cambridge\\
  Cambridge, UK\\
  \texttt{mj201@cam.ac.uk} \\
  \And
  Pietro Li\`{o} \\
  University of Cambridge\\
  Cambridge, UK\\
  \texttt{pl219@cam.ac.uk} \\
}

\begin{document}



\maketitle

\begin{abstract}
The opaque reasoning of Graph Neural Networks induces a lack of human trust. Existing graph network explainers attempt to address this issue by providing post-hoc explanations, however, they fail to make the model itself more interpretable. To fill this gap, we introduce the Concept Encoder Module, the first differentiable concept-discovery approach for graph networks. The proposed approach makes graph networks explainable by design by first discovering graph concepts and then using these to solve the task. Our results demonstrate that this approach allows graph networks to: (i) attain model accuracy comparable with their equivalent vanilla versions, (ii) discover meaningful concepts that achieve high concept completeness and purity scores, (iii) provide high-quality concept-based logic explanations for their prediction, and (iv) support effective interventions at test time: these can increase human trust as well as significantly improve model performance.

\end{abstract}

\section{Introduction}
Human trust in machine learning requires high task performance alongside interpretable decision making~\citep{shen2022trust}. For this reason, the opaqueness of Graph Neural Networks (GNNs,~\citep{scarselli2008graph})---despite their state-of-the-art performance~\citep{battaglia2016interaction,pal2020pinnersage,stokes2020deep,davies2021advancing}---raises ethical~\citep{duran2021afraid, lo2020ethical} and legal~\citep{wachter2017counterfactual, gdpr2017} concerns. As their practical deployment is now under question, interpreting GNN reasoning has become a major concern in the field~\citep{rudin2019stop,ying2019gnnexplainer}.

Early explainability methods for GNNs produce local, low-level post-hoc explanations~\citep{ying2019gnnexplainer,luo2020parameterized,vu2020pgm}, which exhibit the same unreliability as analogous methods for convolutional networks~\citep{adebayo2018sanity, kindermans2019reliability, ghorbani2019interpretation}. In contrast, concept-based explainability overcomes the brittleness of low-level explanations by providing robust global explanations in form of human-understandable concepts~\citep{kim2018interpretability}, i.e., interpretable high-level units of information~\citep{ghorbani2019towards,koh2020concept}. In relational learning, the Graph Concept Explainer (GCExplainer,~\citep{magister2021gcexplainer}) pioneered concept-based explainability for GNNs by extracting global subgraph-based concepts, such as a ``house-shaped'' structure, from the latent space of a trained model.
This way users can check whether the extracted concepts are meaningful~\citep{ghorbani2019towards}, whether they are coherent across samples~\citep{magister2021gcexplainer}, and whether they contain sufficient information to solve a target task~\citep{yeh2020completeness}. However, as any post-hoc approach, GCExplainer does not encourage the GNN to make interpretable predictions using the extracted concepts~\citep{rudin2019stop}. As a result, the opaque reasoning of GNNs ultimately remains an open problem.

\begin{figure}[!t]
    \centering
    \includegraphics[width=0.8\textwidth]{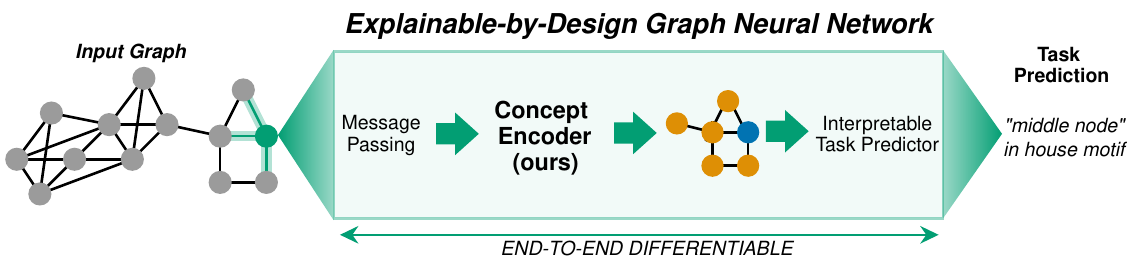}
    \caption{The proposed concept encoder module makes graph neural networks explainable-by-design by first discovering a set of concepts and then using these to solve the task with an interpretable classifier.}
    \label{fig:abstract}
\end{figure}

To fill this knowledge gap, we propose the Concept Encoder Module (CEM, Figure~\ref{fig:abstract}), the first concept-based end-to-end differentiable approach which makes graph networks \textbf{explainable by design}. It achieves this by first discovering a set of concepts and then using these to solve the task at hand. Our module can be introduced in any GNN architecture. We will refer to the resulting family of architectures as ``Concept Graph Networks'', or CGNs. We experimentally show that CGNs: (i) attain better or competitive task accuracy w.r.t. their equivalent vanilla GNN, (ii) encode coherent human-understandable concepts and obtain high scores in all the key concept-based explainability metrics, i.e., purity and completeness, (iii) can provide simple and accurate logic explanations based on discovered concepts, (iv) allow effective interventions at concept level: these can increase human trust and significantly improve model performance.

\section{Background}

\paragraph{Graph Neural Networks}
Graph Neural Networks (GNNs,~\citep{scarselli2008graph}) are differentiable models designed to process relational data in the form of graphs. A graph can be defined as a tuple $G=(V,E)$ which comprises nodes $V = \{1, \dots, n\}$, the entities of a domain, and edges $E \subseteq \{1, \dots, n\} \times \{1, \dots, n\}$, the relations between pairs of nodes. Nodes (or edges) can be endowed with features $\mathbf{x}_i \in \mathbb{R}^d$, representing $d$ characteristics of each entity (or relation), and with $l$ ground truth task labels $y_i \in Y \subseteq \{0, 1\}^l$. A typical GNN $g$ learns a set of node embeddings $\mathbf{h}_i$ with a scheme known as message passing~\citep{gilmer2017neural}. Specifically, message passing aggregates for each node $i\in V$ local information shared by its neighboring nodes $N_i = \{k: (k,i) \in E \}$:
\begin{equation}
    \mathbf{h}_i = \sum_{k \in N_i} g(\mathbf{m}_{ik}, \mathbf{x}_i) \qquad \mathbf{m}_{ik} = \text{msg}(\mathbf{x}_i, \mathbf{x}_k)
\end{equation}
A readout function $f: H \rightarrow Y$ then processes the node embeddings to predict node labels $\hat{y}$. GNNs are trained via stochastic gradient descent minimizing the cross entropy loss between predicted $\hat{y}_i$ and ground-truth task labels $y_i$.





\paragraph{Graph Concept Explainer}
The Graph Concept Explainer (GCExplainer,~\citep{magister2021gcexplainer}) is the first (and only) concept-based approach for interpreting GNNs. Following methods successfully applied in vision~\citep{ghorbani2019towards}, GCExplainer is an unsupervised approach for post-hoc discovery of global concepts. It achieves this by applying k-Means clustering~\citep{forgy1965cluster} on the node embeddings $\mathbf{h}_i$ of a trained GNN. \citet{magister2021gcexplainer} argued that each of the $k$ clusters possibly represents a learnt concept according to human perception, as already suggested by~\citet{zhang2018unreasonable} and~\citet{yeh2020completeness}. Using this clustering, GCExplainer then assigns a concept label $\hat{c}_i \in \hat{C} \subseteq \{0,1\}^k$ to each sample. Finally, it represents each concept using the five samples closest to each cluster centroid, where each sample is visualized as a subgraph with the corresponding node and its $p$-hop neighbors. This visualization technique is effective and aligned with the reasoning of GNNs, as it takes into account how the information flows via message passing. For example, after three layers of message passing, each node can receive at most information from its $3$-hop neighbors.

\paragraph{Trust through concepts and interventions}
Predicting tasks as a function of learnt concepts makes the decision process of deep learning models more interpretable~\citep{koh2020concept, shen2022trust}.
In fact, learning intermediate concepts allows models to provide concept-based explanations for their predictions~\citep{ghorbani2019interpretation} which can take the form of simple logic statements, as shown by~\citet{ciravegna2021logic}. In addition, \citet{koh2020concept} showed how learning intermediate concepts allows human experts to rectify mispredicted concepts through effective test-time interventions, thus improving model's performance and engendering human trust~\citep{shen2022trust}.

\section{Encoding Concepts in Graph Neural Networks} \label{sec:method}
GCExplainer significantly improves the state of the art in GNN explainability, paving the way for concept-based approaches~\citep{magister2021gcexplainer}. However, all existing methods for graph explainability, including GCExplainer, are post-hoc~\citep{ying2019gnnexplainer,luo2020parameterized,vu2020pgm, bajaj2021robust, lucic2022cf}. At best, post-hoc techniques can correctly describe what models learn~\citep{rudin2019stop}, but they cannot make the GNN itself more interpretable. Therefore, the opaque reasoning of GNNs remains an open problem. To fill this knowledge gap, we propose the \textit{Concept Encoder Module} (CEM), a differentiable approach which makes GNNs explainable by design. In fact, CEM empowers GNNs through a more interpretable decision making process. It achieves this by first extracting a set of concepts and then using these to solve a classification task. Our approach can be integrated in any GNN architecture. We will refer to the resulting family of architectures as ``Concept Graph Networks'', or CGNs. As in GCExplainer, humans can visualize CGN concepts to check whether they are meaningful and coherent. Yet, in contrast to GCExplainer, CGNs allow effective interventions at concept level, allowing human experts to significantly improve model performance. CEM integrates a differentiable concept encoding layer to extract node-level and graph-level concepts with an interpretable task predictor providing logic-based explanations. 



\paragraph{Concept encoding}
The first CEM step consists in extracting node-level clusters corresponding to concepts. However, in contrast to GCExplainer, in CEM this step is differentiable and integrated in the network architecture, allowing gradients to optimize clusters in GNN embeddings. Specifically, we implemented this differentiable clustering using a normalized softmax activation on the node-level embeddings $\mathbf{h}_i$, associating each node with one cluster/concept. This operation returns for each node a fuzzy encoding $\mathbf{q}_i \in [0,1]^m$:
\begin{equation} \label{eq:diffGCExp}
    \tilde{\mathbf{q}}_i = \frac{\exp({\mathbf{h}_i})}{\sum_{u=1}^m \exp(\mathbf{h}_{iu})}, \qquad \mathbf{q}_i = \frac{\tilde{\mathbf{q}}_i}{\max_i \tilde{\mathbf{q}}_i + \epsilon}
\end{equation}
CEM then clusters nodes considering the similarity of their fuzzy encodings $\mathbf{q}_i$. Specifically, CEM groups the samples together depending on their Booleanized encoding $\mathbf{r}_i \in \{0,1\}^m$:
\begin{equation}
    \mathbf{r}_{iu} = 
    \begin{cases}
    1 \quad \text{ if } \mathbf{q}_{iu} \geq \tau\\
    0 \quad \text{ otherwise }
    \end{cases}
\end{equation}
where $\tau \in [0,1]$ is conventionally set to $0.5$.
In particular, two samples $a$ and $b$ belong to the same cluster if and only if their encodings $\mathbf{r}_a$ and $\mathbf{r}_b$ match.  For example, consider the two node embeddings $\mathbf{h}_a = [-1.2, 2.3]$ and $\mathbf{h}_b = [2.2, 1.8]$. For these inputs, the normalized softmax would return the fuzzy encodings $\mathbf{q}_a = [0.0293, 0.9707]$ and $\mathbf{q}_b = [0.5987, 0.4013]$, respectively. As their Booleanizations $\mathbf{r}_a = [0, 1]$ and $\mathbf{r}_b = [1, 0]$ do not match, we can then conclude that the two nodes belong to different clusters. Notice how our concept encoding is highly efficient, as it allows to learn up to $2^m$ different concepts on GNN embeddings $\mathbf{h}_i$ of size $m$. This way the GNN can dynamically find the optimal number of concepts/clusters, thus relieving users from this burden. In fact, users just need to choose an upper bound to the number of concepts $k$ rather than an exact value as in k-Means for GCExplainer. In order to account for graph classification, the concept encodings for a graph are pooled before being passed to the interpretable model predicting the task, as explained in the next paragraph.

\paragraph{Interpretable predictions}
The second CEM step consists of using the extracted concepts to make interpretable predictions for downstream tasks. In particular, the presence of concepts enables pairing GNNs with existing concept-based methods which are explainable by design, such as Logic Explained Networks (LENs,~\citep{ciravegna2021logic}). LENs are neural models providing simple concept-based logic explanations for their predictions. Specifically, LENs can provide class-level explanations which makes our approach the first at providing unique global explanations for GNNs. CEM uses a LEN as the readout function $f$ for the classification, applying it on top of concept representations $\mathbf{q}_i$. For graph classification tasks, the input data is composed of a set of $t$ graphs $G^j \in \{(V^j, E^j)\}_{j=1}^t$, where each graph is associated with a task label $y^j \in Y$. In this setting, GNN-based models predict a single label for each graph $G^j$ by pooling its node-level encodings $\mathbf{q}_i^j$ to aggregate over multiple concepts:
\begin{equation} \label{eq:lens}
    \hat{y}_i = \text{LEN}_{\text{node}} ( \mathbf{q}_i), \qquad
    \hat{y}^j = \text{LEN}_{\text{graph}} \Bigg(\frac{1}{n_j} \sum_{i=1}^{n_j} \mathbf{q}_i^j \Bigg)
\end{equation}
where $n_j$ is the number of nodes associated with graph $j$. In our implementation, we use the entropy-based layer to implement LENs~\citep{barbiero2021entropy}) as it can provide high classification accuracy with high-quality logic explanations. This entropy-based layer implements a sparse attention layer designed to work on top of concept activations.
The attention mechanism allows the model to focus on a small subset of concepts to solve each task. It also introduces a parsimony principle in the architecture corresponding to an intuitive human cognitive bias~\citep{miller1956magical}. This parsimony principle allows the extraction of simple logic explanations from the network, thus making these models explainable by design. 


\paragraph{Concept-based and logic-based explanations}
The proposed method provides two types of explanations: concept-based and logic-based explanations. Global concept-based explanations can be extracted in a similar manner as in GCExplainer: a concept for a node or graph is extracted by finding the cluster with which a node's embedding is associated, and visualising the samples closest to the cluster centroid. The logic-based formula provided per class broadens the explanation scope, as it indicates which neurons of the concept encoding $\mathbf{q}_i$ are activated and representative of a class. This provides a more comprehensive explanation since a class can be associated with multiple concepts. 

\paragraph{Concept interventions}
As in Concept Bottleneck Models~\citep{koh2020concept}, our approach supports human interaction at concept level. In fact, in contrast to existing post-hoc methods, a explainable-by-design approach creates an explicit concept layer which can positively react to test-time human interventions. For instance, consider a misclassified node with concept encoding $\mathbf{q}_a = [0.21, 0.93]$. Assume that the vast majority of nodes with the binary encoding $\mathbf{r}_{\text{grid\_node}} = [0, 1]$ are nodes of a grid-like structure, which allows a human to label this cluster as ``grid nodes''. Now, a human expert can inspect the neighborhood of the misclassified node and realize that this node belongs to a circle-like structure and not to a grid structure. As the binary encoding for the concept ``circle nodes'' is $\mathbf{r}_{\text{circle\_node}} = [1, 1]$, the user can easily apply an intervention to correct the misclassified concept by changing its encoding to $\mathbf{q}_a:=[1, 1]$. Such an update allows the interpretable readout function to act on information related to the corrected concept, thus improving the original model prediction.

\section{Experiments} \label{sec:experiments}
In our experiments we focus on the following research questions:
\begin{itemize}
    \item \textbf{Task accuracy and completeness ---} What is the impact of our approach on the generalization error of a GNN? Is the identified concept set complete w.r.t. the task?
    \item \textbf{Concept interpretability ---} Are the unsupervised concepts identified by our model meaningful? Do they match ground truths or human expectations?
    \item \textbf{Explanation performance ---} Are concepts pure and coherent? Are the logic explanations provided by our model accurate? Are they simple enough to be interpretable?
\end{itemize}
With these questions in mind, we hypothesize that our approach can: (i) obtain similar task accuracy w.r.t. a standard GNN; (ii) extract the ground truth graph concepts aligned with human expectations, and (iii) identify pure concepts as well as simple and accurate logic explanations.

\paragraph{Metrics}
In our evaluation, we measure model performance and interpretability based on five metrics. We measure model performance via \textit{classification accuracy} to compare the generalization error of CGNs w.r.t. their equivalent vanilla GNNs. To evaluate model interpretability, we compute \textit{concept completeness}~\citep{yeh2020completeness}, which assesses whether the concepts discovered are sufficient to describe the downstream task. Following~\citet{yeh2020completeness}, we use a decision tree~\citep{breiman1984classification} to predict the task labels given the concept encoding associated with each input instance. We also examine concept coherence via \textit{concept purity}~\citep{magister2021gcexplainer}. Following~\citet{magister2021gcexplainer}, we measure concept purity by considering the graph edit distance of samples' neighborhoods within each cluster/concept. Having checked concept quality, we evaluate logic explanations in terms of their accuracy and complexity. We calculate the \textit{accuracy of logic explanations} using the learnt logic formulas for classifying test samples based on their concept encoding as done by~\citet{ciravegna2021logic}. This mirrors the computation of concept completeness, however, instead of a decision tree, we use the learnt logic formulas for classification. Lastly, we evaluate the \textit{complexity of logic explanations} by measuring the number of terms in logic rules~\citep{ciravegna2021logic}. We compute all metrics on test sets across five random weight initializations and report their means and $95\%$ confidence intervals using the Box-Cox transformation for non-normal distributions. Further details on our evaluation can be found in the Appendix. We do not measure classical explainability metrics, such as sensitivity and sparsity~\citep{zhou2021evaluating}, as they apply to explainers of models rather than explainable-by-design networks themselves.
    
\paragraph{Datasets}
We perform the experiments on the same set of datasets as GNNExplainer~\citep{ying2019gnnexplainer}, as subsequent research established them as benchmarks~\citep{luo2020parameterized,vu2020pgm,magister2021gcexplainer}. We use five synthetic node classification datasets, which have a ground truth motif encoded. A ground truth motif is a subgraph in the graph, which a successful GNN explainability technique should recognize. The node classification datasets are: (i) BA-Shapes, consisting of a single graph where the base structure is a Barab\'asi-Albert (BA) graph~\citep{barabasi1999emergence}, which has $80$ house motifs and $70$ random edges attached to it; (ii) BA-Community, generated by the union of two BA-Shapes graphs with the task of classifying nodes belonging to house motifs, as well as the membership of the node in one of the two houses; (iii) BA-Grid, based on a grid motif attached to a BA graph, similarly to BA-Shapes; (iv) Tree-Cycles, formed by a binary tree with $80$ cycle structures attached; (v) Tree-Grid, consisting of a binary tree with $80$ grid structures attached. We also include two real-world datasets to evaluate model performance on less structured data and on graph classification tasks. The two real world datasets are: (i) Mutagenicity (MUTAG,~\citep{morris2020tudataset}), a collection of graphs representing molecules labelled as mutagenic or non-mutagenic, and (ii) Reddit-Binary (Reddit,~\citep{morris2020tudataset}), a collection of graphs representing the structure of Reddit discussion threads where nodes represent users and edges the interaction between users. A challenge in evaluating these datasets is that there are no ground truth motifs. However,~\citet{ying2019gnnexplainer} suggests the ring structure and the nitrogen dioxide compound in Mutagenicity, and the star-like structure in Reddit-Binary as desirable motifs to be recovered.

\paragraph{Baselines and setup}
To address our research questions, we compare our approach against an equivalent convolutional vanilla GNN explained by GCExplainer. Notice that we do not focus on other post-hoc explainability methods, such as GNNExplainer~\citep{ying2019gnnexplainer}, PGExplainer~\citep{luo2020parameterized} or PGM-Explainer~\citep{vu2020pgm}, as to the best of our knowledge GCExplainer is the only explainability method providing global concept-based explanations for GNNs. We refer the reader to the Appendix for a comparison of the proposed method with GNNExplainer and for an evaluation of the proposed approach with further GNN architectures.

We select the models' hyperparameters, such as the number of hidden units and learning rate, using a grid search. To ensure fairness in our results, we use the same architecture capacity and hyperparameters for our model as well as for its vanilla counterpart. We initialize the hyperparameters of GCExplainer to the values determined experimentally by~\citet{magister2021gcexplainer}. We refer the reader to the Appendix for a detailed discussion of the setup.

\section{Results} \label{sec:results}

\subsection{Task Accuracy and Completeness}
\begin{figure}[!t]
    \centering
    \includegraphics[width=\textwidth]{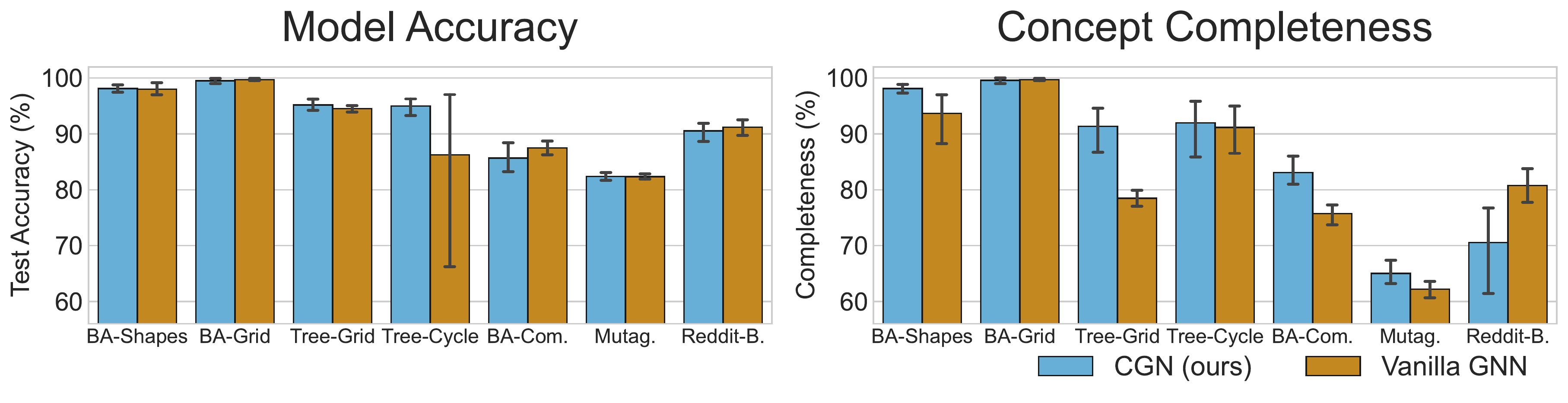}
    \caption{Model accuracy and concept completeness for the Concept-based Graph Network (CGN) and an equivalent vanilla GNN. For these results, and those that follow, we compute all metrics on test sets across five seeds and report their mean and $95\%$ confidence intervals.}
    \label{fig:accuracy}
\end{figure}

\paragraph{Concept Graph Networks are as accurate as vanilla GNNs (Figure \ref{fig:accuracy} left)}
Our results show that CEM allows GNNs to achieve better or comparable task accuracy w.r.t. equivalent GNN architectures. Specifically, our approach outperforms vanilla GNNs on the Tree-Cycle dataset, having a higher test accuracy (plus $\sim 8\%$ on average) and less variance across different parameter initializations. We hypothesize that this effect is due to more stable and pure concepts being learnt thanks to CEM, as we will see later when discussing Figure~\ref{fig:purity}. We do not observe any significant negative effect of using CEM on the generalization error of GNNs.

\paragraph{The Concept Encoder Module discovers complete concepts (Figure \ref{fig:accuracy} right)}
Our experiments show that overall CEM discovers a more complete set of concepts w.r.t. the concept set extracted by GCExplainer on equivalent GNN architectures. This is particularly emphasized in the Tree-Grid, BA-Shapes, BA-Community, and Mutagenicity datasets, where CEM significantly outperforms GCExplainer by up to $\sim 12\%$. For the other datasets, the proposed approach matches the concept completeness scores of GCExplainer, except for the Reddit-Binary dataset. However, a detailed analysis of completeness scores across the five different runs reveals that the mean is skewed by a single run. In absolute terms, CEM discovers highly complete sets of concepts with completeness scores close to the model accuracy. 

\subsection{Concept Interpretability}
\paragraph{The Concept Encoder Module identifies meaningful concepts (Table~\ref{tab:concept_visuals})}
CEM discovers high-quality concepts, which are meaningful to humans. Similarly to GCExplainer, our results demonstrate that CEM can discover concepts corresponding to the ground truth motifs embedded in the toy datasets. For example, our approach recovers the ``house motif'' in BA-Shapes. Moreover, CEM proposes plausible concepts for the real-world datasets where ground truth motifs are lacking. In this case, the extracted concepts match the desirable motifs suggested by~\citet{ying2019gnnexplainer}, corresponding to ring structures and the nitrogen dioxide compound in Mutagenicity, and a star-like structure in Reddit-Binary. 
As we use the same visualization technique as GCExplainer the merit of our contribution lies in the discovery of a more descriptive set of concepts, which includes rare and fine-grained concepts. As a thorough qualitative comparison with GCExplainer requires exhaustive visualization, we refer the reader to the Appendix for a complete set of results.



\newcommand\conceptsizef{40}
\newcommand\vfigsf{-.5\height}

\begin{table}[!t]
\centering
\renewcommand{\arraystretch}{1}
\resizebox{\textwidth}{!}{%
\begin{tabular}{cccccccc}
\toprule
 & \textbf{BA-Shapes} & \textbf{BA-Grid} & \textbf{Tree-Grid} & \textbf{Tree-Cycle} & \textbf{BA-Community} & \textbf{Mutagenicity} & \textbf{Reddit-Binary} \\
\midrule
\textbf{Ground Truth} & \raisebox{\vfigsf}{\includegraphics[height=\conceptsizef pt]{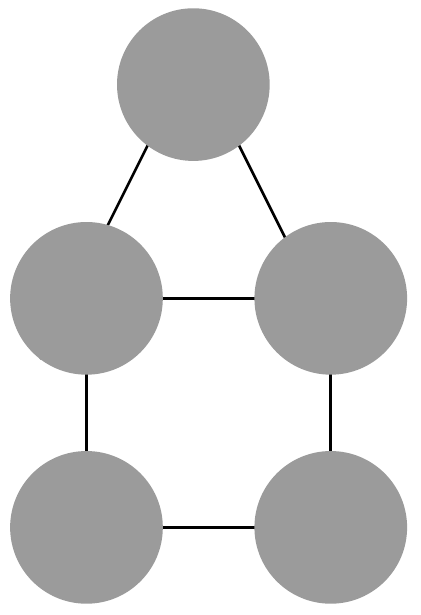}} & \raisebox{\vfigsf}{\includegraphics[height=\conceptsizef pt]{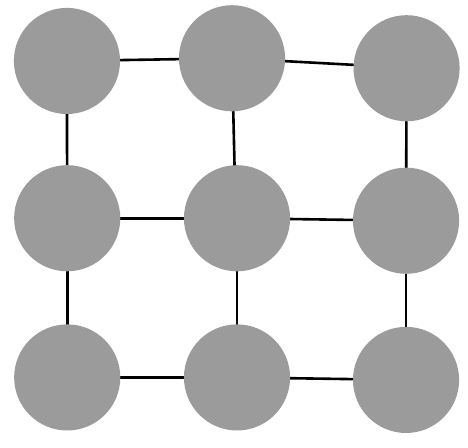}} & \raisebox{\vfigsf}{\includegraphics[height=\conceptsizef pt]{fig/logic_expl_graphs/grid.pdf}} & \raisebox{\vfigsf}{\includegraphics[height=\conceptsizef pt]{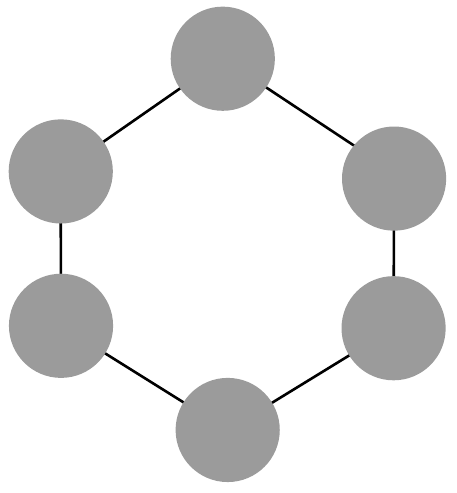}} & \raisebox{\vfigsf}{\includegraphics[height=\conceptsizef pt]{fig/logic_expl_graphs/house.pdf}} & \raisebox{\vfigsf}{\includegraphics[height=\conceptsizef pt]{fig/logic_expl_graphs/ring.pdf}}  \raisebox{\vfigsf}{\includegraphics[height=40 pt]{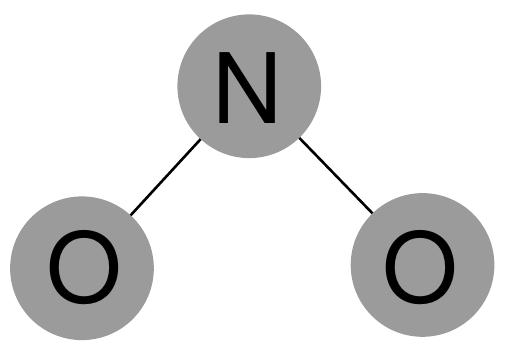}} & \raisebox{\vfigsf}{\includegraphics[height=\conceptsizef pt]{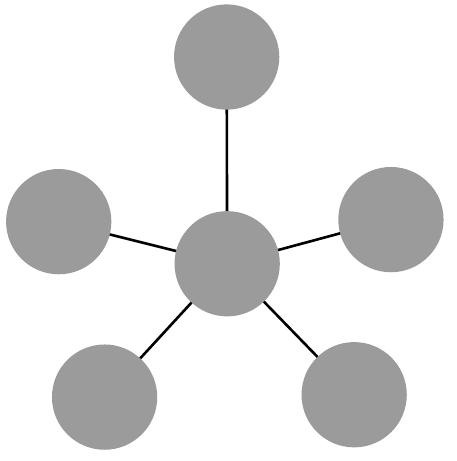}}  \\
\textbf{Extracted Concept} & \raisebox{\vfigsf}{\includegraphics[height=\conceptsizef pt]{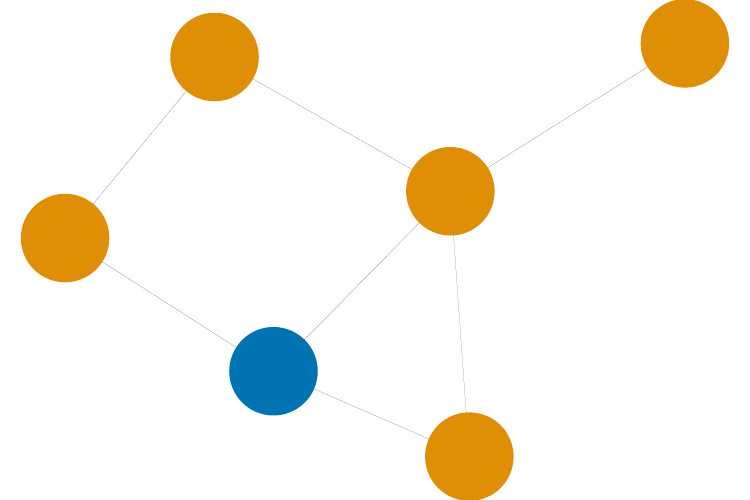}} & \raisebox{\vfigsf}{\includegraphics[height=\conceptsizef pt]{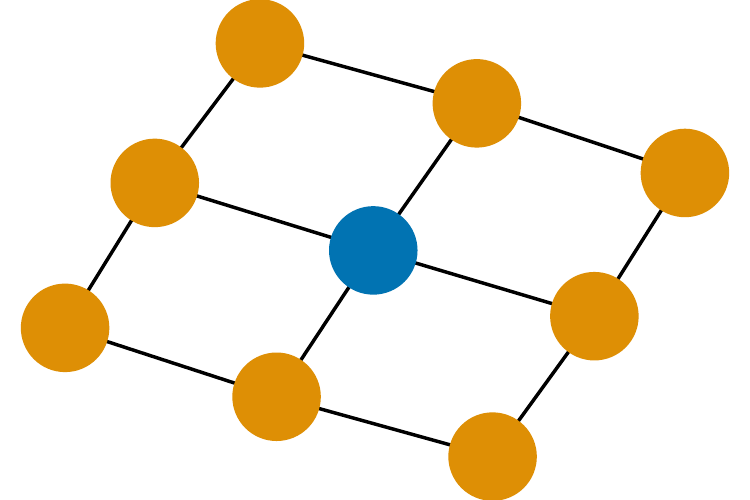}} & \raisebox{\vfigsf}{\includegraphics[height=\conceptsizef pt]{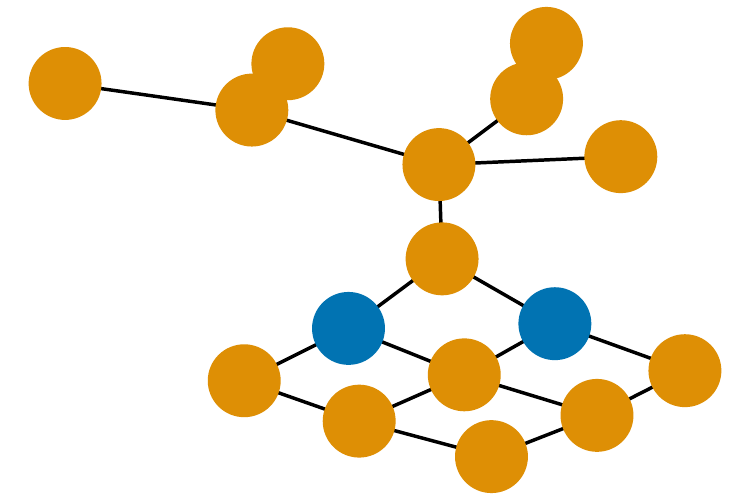}} & \raisebox{\vfigsf}{\includegraphics[height=\conceptsizef pt]{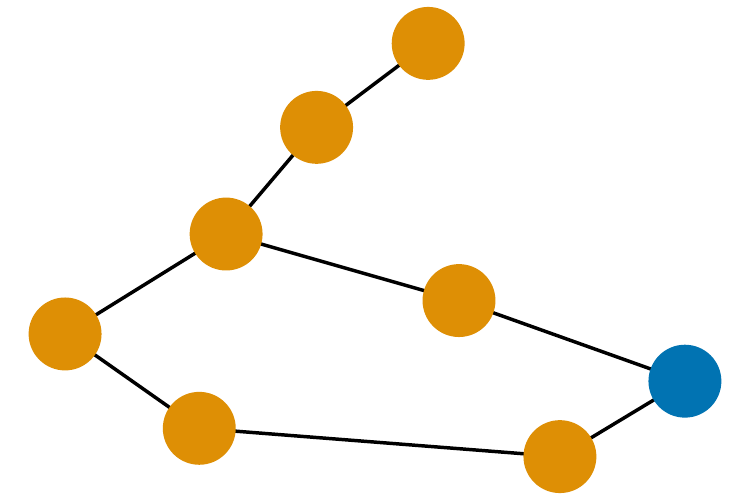}} & \raisebox{\vfigsf}{\includegraphics[height=\conceptsizef pt]{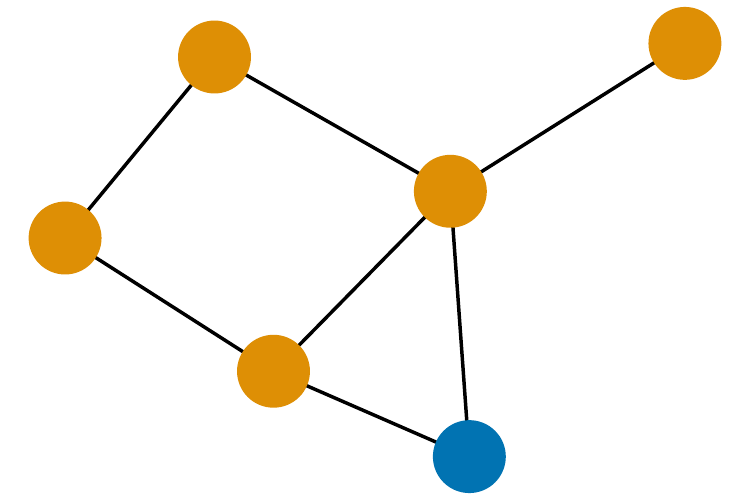}} & \raisebox{\vfigsf}{\includegraphics[height=\conceptsizef pt]{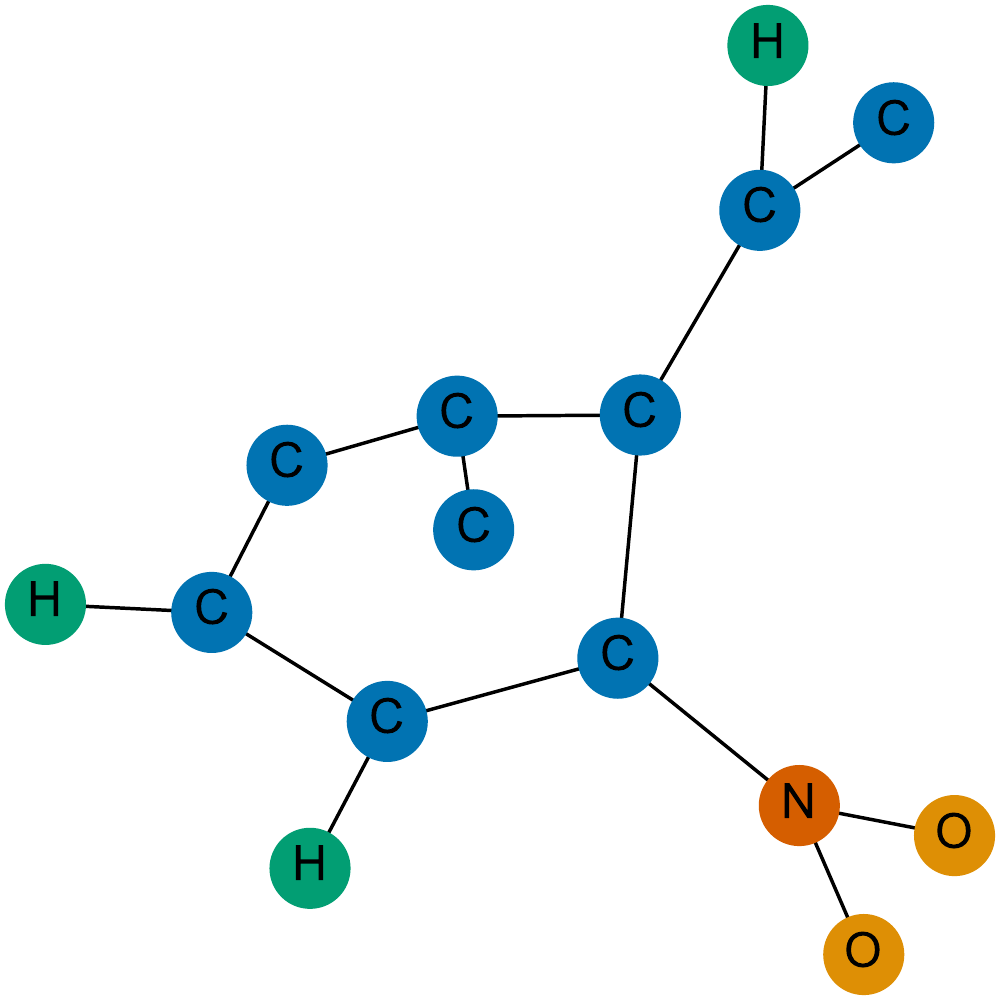}} & \raisebox{\vfigsf}{\includegraphics[height=\conceptsizef pt]{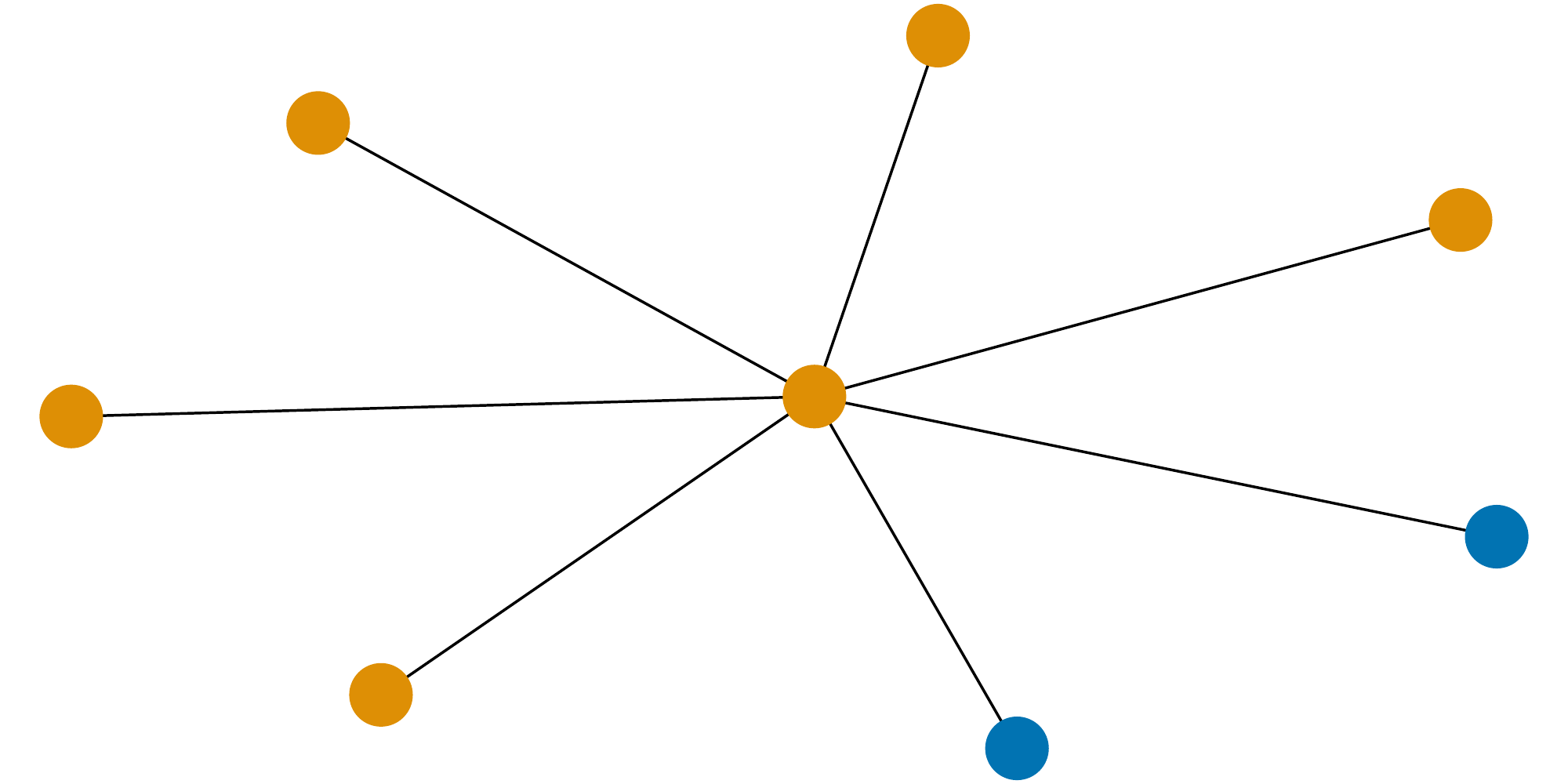}}  \\
\bottomrule\\
\end{tabular}%
}
\caption{The Concept Graph Module detects meaningful concepts matching the expected ground truth. Blue nodes are the instances being explained, while orange nodes represent their $p$-hop neighbors. Similar motifs are identified by GCExplainer.}
\label{tab:concept_visuals}
\end{table}

\begin{table}[!t]
\centering
\renewcommand{\arraystretch}{1}
\resizebox{0.85\textwidth}{!}{%
\begin{tabular}{ccccc}
\toprule
\textbf{Ground Truth} & \multicolumn{2}{c}{\textbf{Fine-Grained Concepts}} & \multicolumn{2}{c}{\textbf{Rare Concepts}} \\
\midrule
\raisebox{\vfigsf}{\includegraphics[height=\conceptsizef pt]{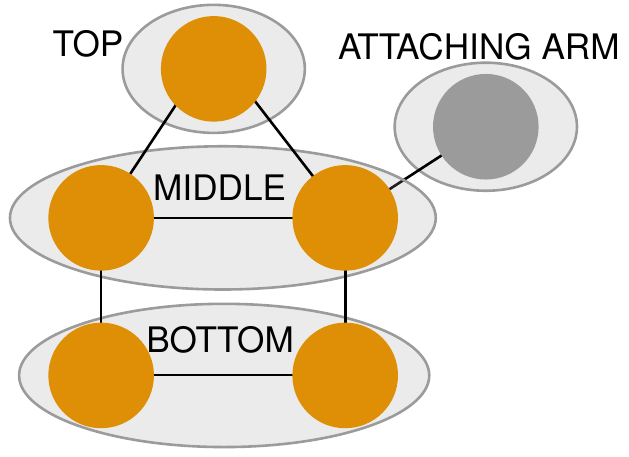}} & \raisebox{\vfigsf}{\includegraphics[height=\conceptsizef pt]{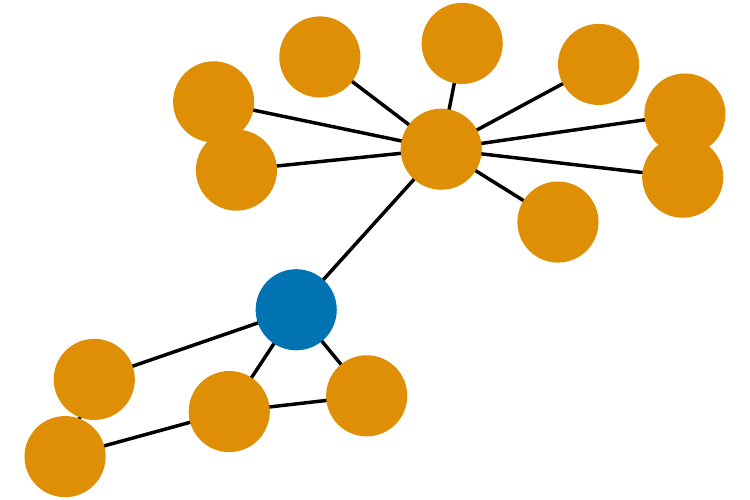}} & \raisebox{\vfigsf}{\includegraphics[height=\conceptsizef pt]{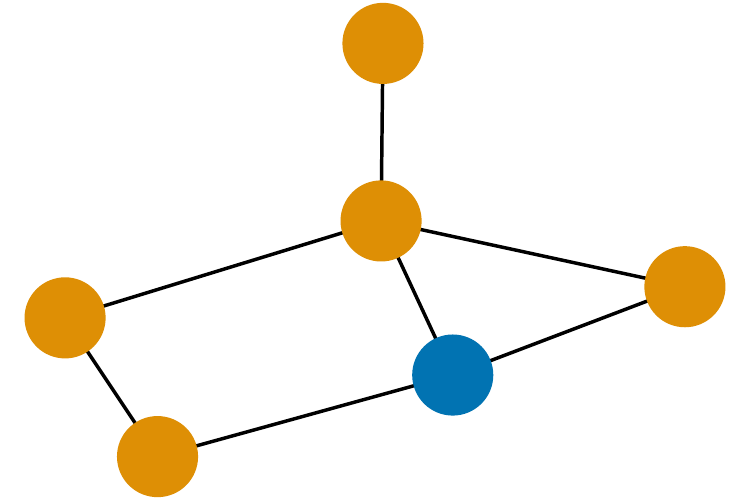}} & \raisebox{\vfigsf}{\includegraphics[height=\conceptsizef pt]{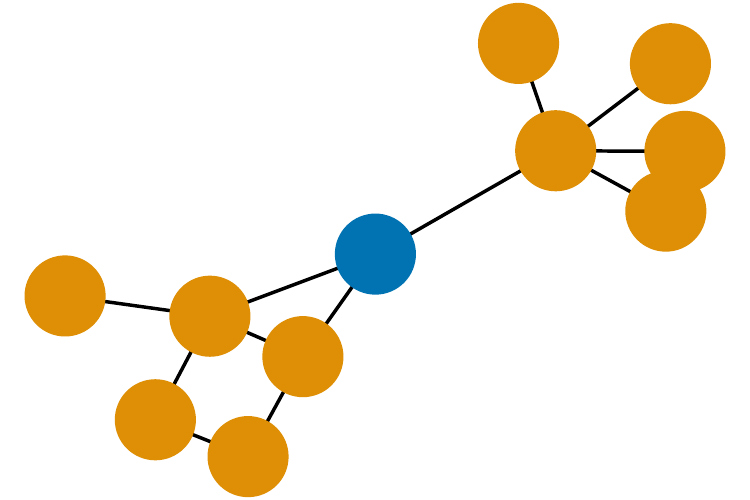}} & \raisebox{\vfigsf}{\includegraphics[height=\conceptsizef pt]{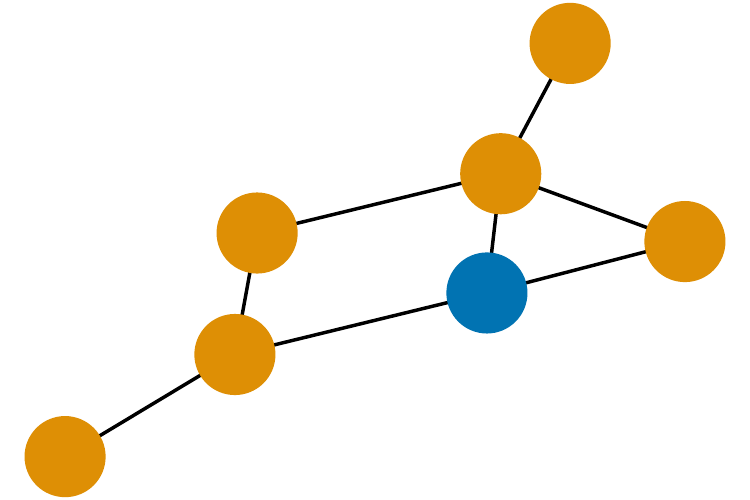}} \\
\bottomrule\\
\end{tabular}%
}
\caption{The Concept Graph Module detects concepts more fine-grained than the simple ground truth motif encoded as well as rare motifs. Blue nodes are the instances being explained, while orange nodes represent their $p$-hop neighbors. Notably, GCExplainer gives no indication of rare concepts.}
\label{tab:rare_concepts}
\end{table}



\paragraph{The Concept Encoder Module identifies rare and fine-grained concepts (Table~\ref{tab:rare_concepts})}
CEM discovers more fine-grained concepts than just the ``house motif'' suggested by GNNExplainer, as it can differentiate whether a middle or bottom node is on the far or near side of the edge attaching to the BA graph. This matches the quality of concepts extracted by GCExplainer. In contrast to GCExplainer, CEM also identifies rare concepts. Rare motifs are present in toy datasets through the insertion of random edges. As the proposed approach can find the optimal number of clusters/concepts dynamically, clusters of a very small size possibly represent rare motifs. To check the presence of rare concepts, we visualize the $p$-hop neighbors of nodes found in small clusters. For example, CEM identifies a rare concept represented as a ``house'' structure attached to the BA graph via the top node of the house in the BA-Shapes dataset. This represents a rare concept as it is generated by the insertion of a random edge. We confirm this observations in other toy datasets, such as BA-Community and Tree-Cycle, where motifs with random edges are clearly identified. We have not identified rare concepts in BA-Grid or Tree-Grid, which may be attributed to the random edges being distributed within the base graph, which has a less definite structure. Due to the lack of expert knowledge, we cannot confirm whether the rare motifs found in Mutagenicity and Reddit-Binary align with human expectations.


\subsection{Explanation performance}


\begin{minipage}{0.4\textwidth}
\paragraph{The Concept Encoder Module identifies pure concepts (Figure~\ref{fig:purity})}
CEM discovers high-quality concepts, which are coherent across samples, as measured by concept purity. In terms of purity, our approach discovers concepts with nearly optimal scores in toy datasets, with a graph edit distance close to zero. For these datasets, CEM provides either better or comparable purity scores when compared to GCExplainer. CEM provides slightly worse purity scores in both the Mutagenicity and Reddit-Binary datasets. However, also in this case the absolute purity of CEM is almost optimal.
\end{minipage}
\hspace{0.02\textwidth}
\begin{minipage}{0.55\textwidth}
    \includegraphics[width=\textwidth]{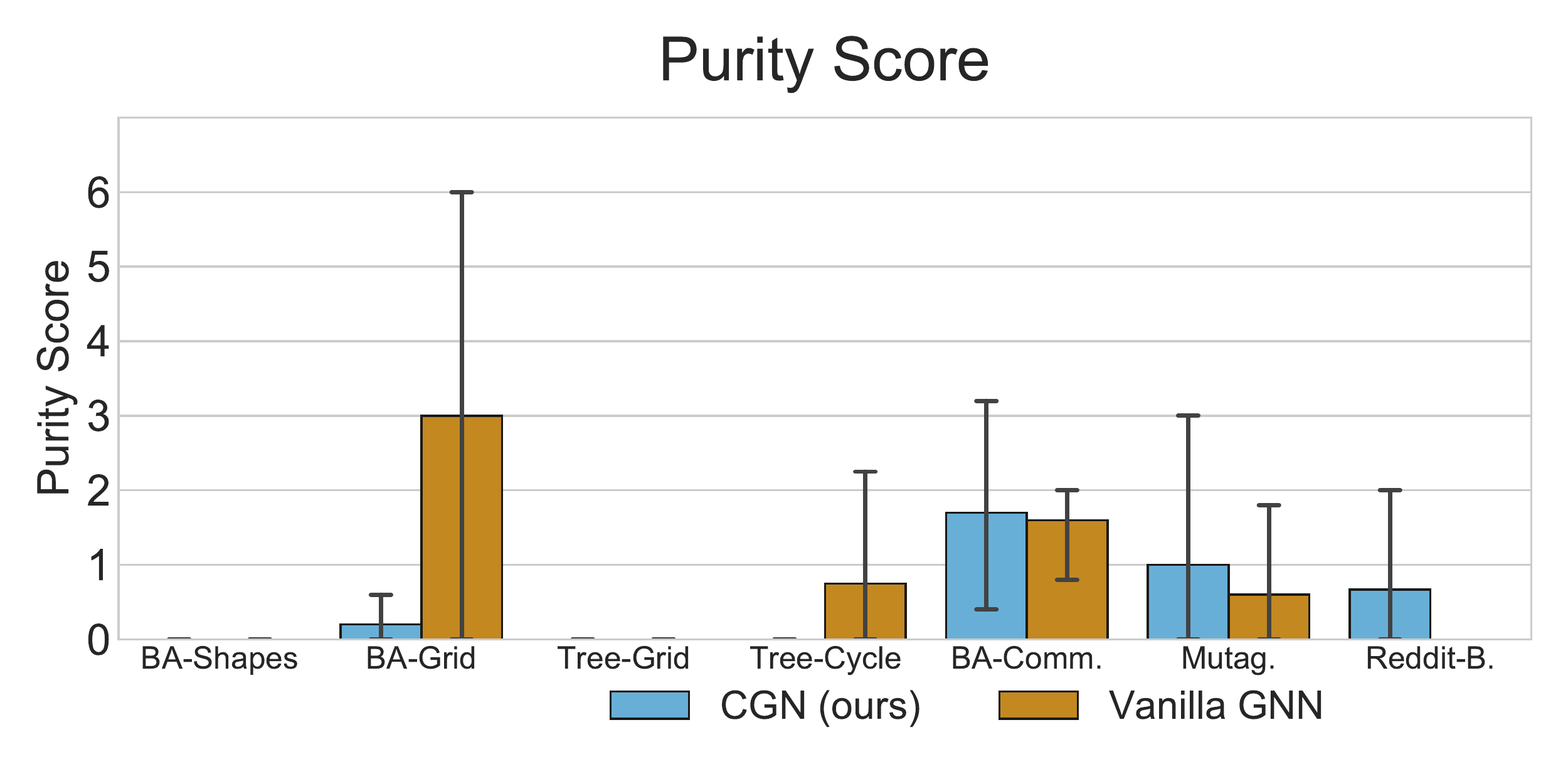}
    \captionof{figure}{Purity scores for the concept extracted by Concept Graph Module and GCExplainer. Notice how the optimal purity score is zero, as it measures the graph edit distance between concept instances~\citep{magister2021gcexplainer}.}
    \label{fig:purity}
\end{minipage}

\begin{figure}[!h]
    \centering
    \includegraphics[width=0.8\textwidth]{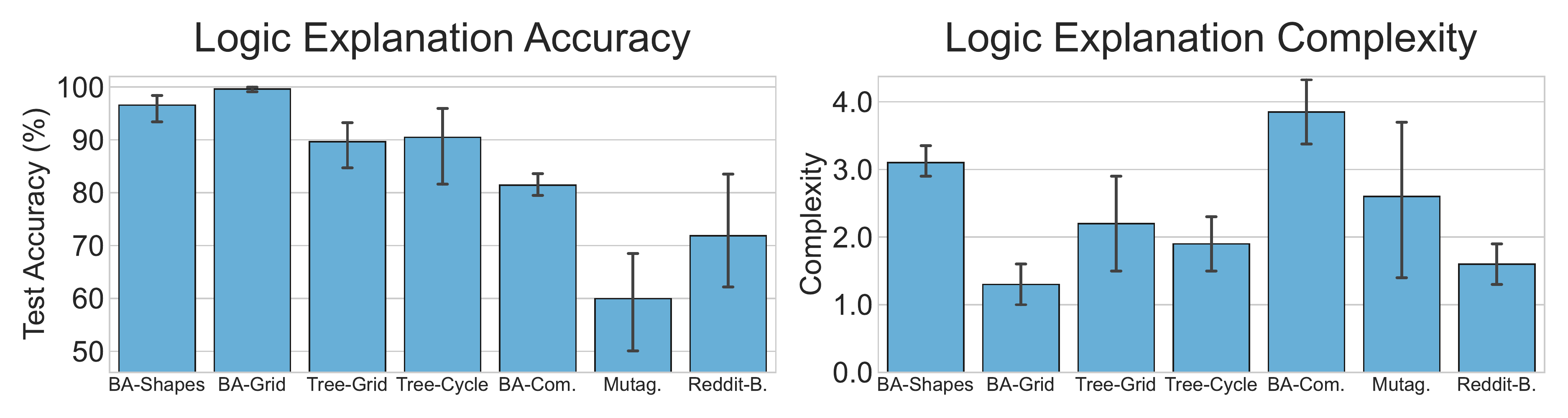}
    \caption{Accuracy and complexity of logic explanations provided by the proposed Concept Graph Module. The accuracy is computed using logic formulas to classify samples based on their concept encoding. Explanation complexity measures the number of minterms in logic formulas.}
    \label{fig:lens}
\end{figure}

\paragraph{The Concept Encoder Module provides accurate logic explanations (Figure \ref{fig:lens}, Table~\ref{tab:logic_explanations})}
LEN allows CEM to provide simple and accurate logic explanations for task predictions. The accuracy of the logic explanations extracted reaches at least $90\%$ across all synthetic datasets, indicating that CEM derives a precise description of the model decision process. Relating the accuracy of explanations back to the model accuracy, we observe that the explanation accuracy is bounded by task performance, as already noticed by~\citet{ciravegna2021logic}. This explains the slightly lower logic explanation accuracy on the real-world datasets, which can be ascribed to the absence of definite ground-truth concepts and to the classification task being more complex. Besides being accurate, logic explanations are very short, with a complexity below $3$ terms. In conjunction with the explanation accuracy, this means that CEM finds a small set of predicates which accurately describes the most relevant concepts for each class. 

\newcommand\conceptsize{40}
\newcommand\vfigs{-.3\height}
\begin{table}[!h]
\centering
\renewcommand{\arraystretch}{1}
\resizebox{\textwidth}{!}{%
\begin{tabular}{llll}
\toprule
\textbf{Dataset} & \textbf{Concept-based Logic Explanation} & \multicolumn{2}{l}{\textbf{Ground Turth Concepts}} \\
\midrule
\textbf{BA-Shapes} & $y =2\leftarrow$ \raisebox{\vfigs}{\includegraphics[height=\conceptsize pt]{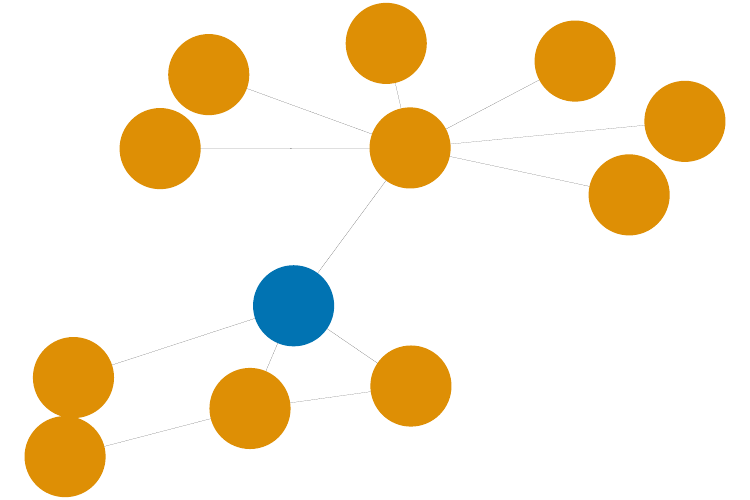}} OR \raisebox{\vfigs}{\includegraphics[height=\conceptsize pt]{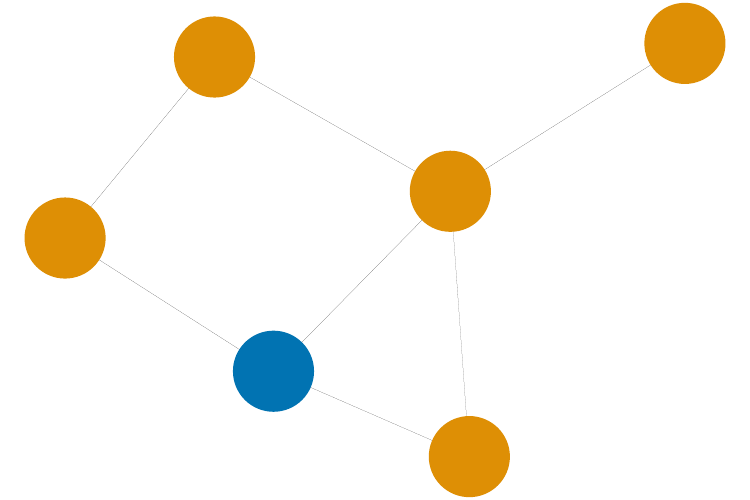}} & \raisebox{\vfigs}{\includegraphics[height=\conceptsize pt]{fig/logic_expl_graphs/house.pdf}}  & \textit{Node in house motif} \\
\textbf{BA-Grid} & $y=1\leftarrow$ \raisebox{\vfigs}{\includegraphics[height=\conceptsize pt]{fig/logic_expl_graphs/BA_Grid_concept_2.pdf}} & \raisebox{\vfigs}{\includegraphics[height=\conceptsize pt]{fig/logic_expl_graphs/grid.pdf}}  & \textit{Node in grid motif}  \\
\textbf{Tree-Grid} & $y=1\leftarrow$ \raisebox{\vfigs}{\includegraphics[height=\conceptsize pt]{fig/logic_expl_graphs/Tree_Grid_concept_21.pdf}} OR \raisebox{\vfigs}{\includegraphics[height=\conceptsize pt]{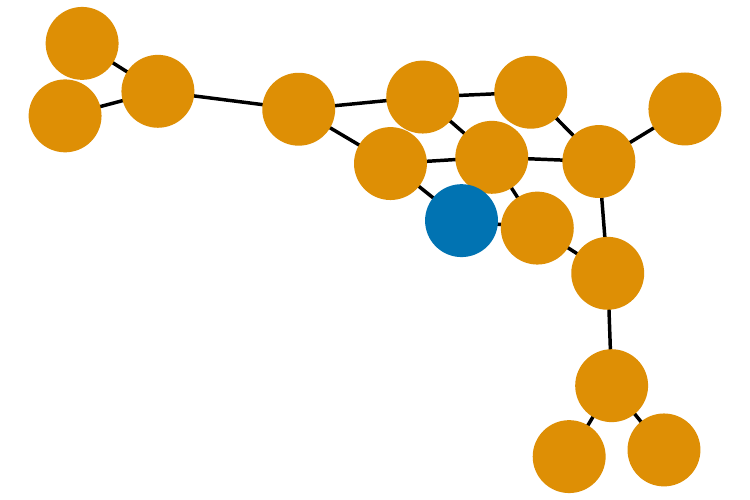}} & \raisebox{\vfigs}{\includegraphics[height=\conceptsize pt]{fig/logic_expl_graphs/grid.pdf}}  & \textit{Node in grid motif} \\
\textbf{Tree-Cycle} & $y=1\leftarrow$ \raisebox{\vfigs}{\includegraphics[height=\conceptsize pt]{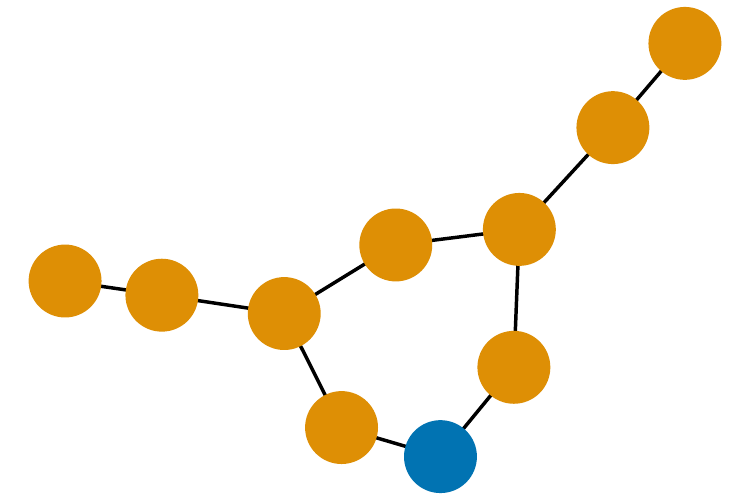}} OR \raisebox{\vfigs}{\includegraphics[height=\conceptsize pt]{fig/logic_expl_graphs/Tree_Cycle_concept_8.pdf}} OR \raisebox{\vfigs}{\includegraphics[height=\conceptsize pt]{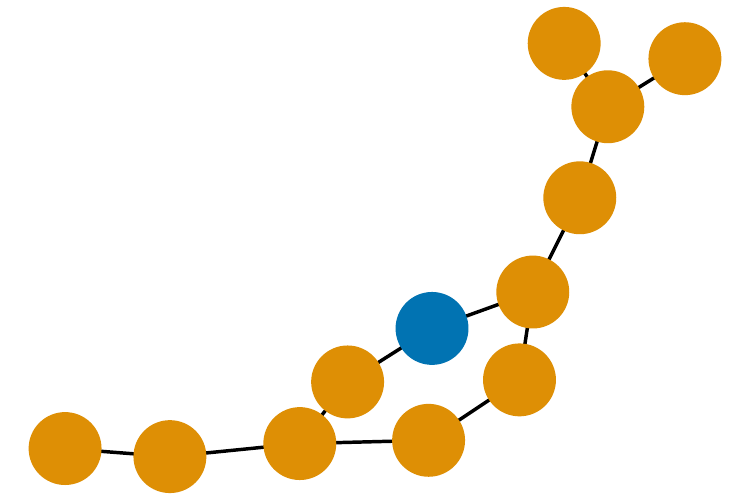}} & \raisebox{\vfigs}{\includegraphics[height=\conceptsize pt]{fig/logic_expl_graphs/ring.pdf}}  & \textit{Node in circle motif} \\
\textbf{BA-Community} & $y=3\leftarrow$ \raisebox{\vfigs}{\includegraphics[height=\conceptsize pt]{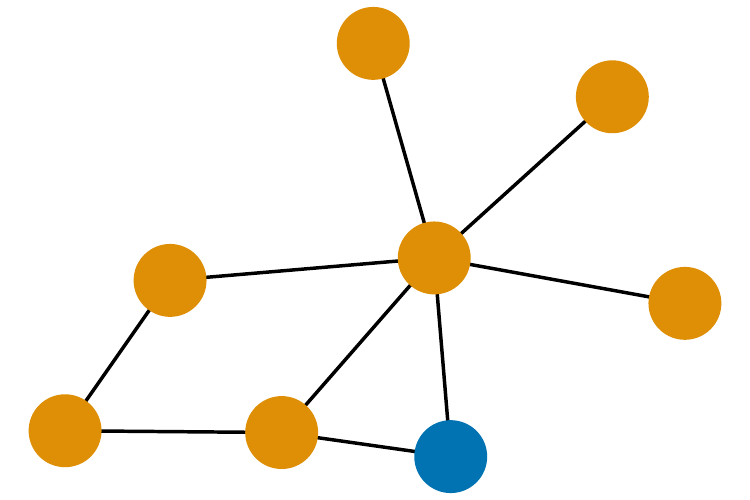}} OR \raisebox{\vfigs}{\includegraphics[height=\conceptsize pt]{fig/logic_expl_graphs/BA_Community_concept_30.pdf}} OR \raisebox{\vfigs}{\includegraphics[height=\conceptsize pt]{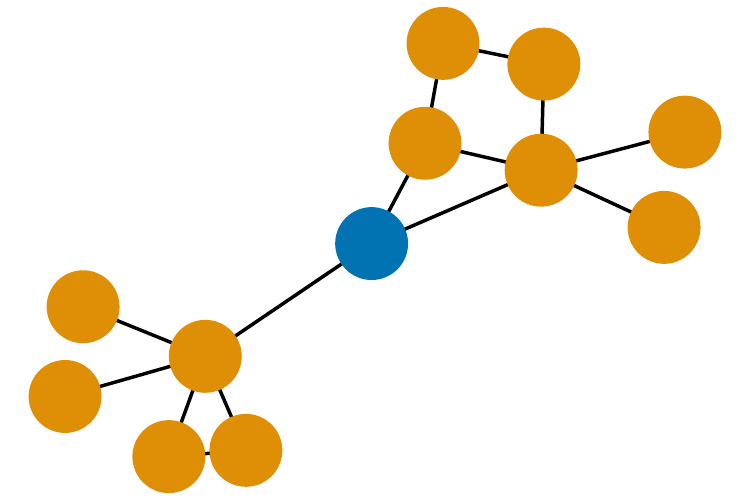}} & \raisebox{\vfigs}{\includegraphics[height=\conceptsize pt]{fig/logic_expl_graphs/house.pdf}}  & \textit{Node in house motif} \\
\textbf{Reddit-Binary} & $y =\text{``Q/A''}\leftarrow$  \raisebox{\vfigs}{\includegraphics[height=\conceptsize  pt]{fig/logic_expl_graphs/Reddit_Binary_concept_24.pdf}} OR \raisebox{\vfigs}{\includegraphics[height=\conceptsize  pt]{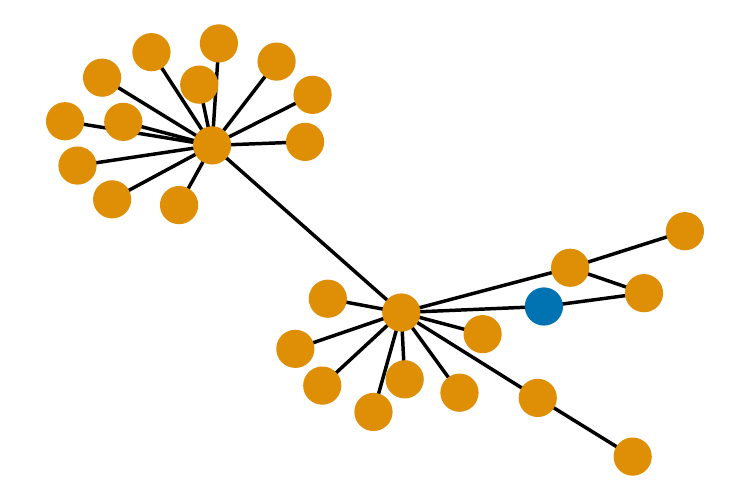}} & \raisebox{\vfigs}{\includegraphics[height=\conceptsize pt]{fig/logic_expl_graphs/star.pdf}}  & \textit{Star motifs} \\
\textbf{Mutagenicity} & $y =\text{``mutagenic''}\leftarrow$ \raisebox{\vfigs}{\includegraphics[height=\conceptsize pt]{fig/logic_expl_graphs/Mutagenicity1.pdf}} & \raisebox{\vfigs}{\includegraphics[height=\conceptsize pt]{fig/logic_expl_graphs/ring.pdf}} \raisebox{\vfigs}{\includegraphics[height=28 pt]{fig/logic_expl_graphs/no2.pdf}}  & \textit{Ring motifs or NO$_{\text{2}}$} \\
\bottomrule\\
\end{tabular}%
}
\caption{An example of a concept-based logic explanations discovered by the Concept Encoder Module for each dataset. Blue nodes are the instances being explained, while orange nodes represent their $p$-hop neighbors. For Mutagenicity the color of each node represents a different chemical element. The logic formulae describe how the presence of concepts can be used to infer task labels. As an example, the first logic rule states that the task label ``middle nodes in house motifs'' ($y=2$) can be inferred from the concepts: ``middle node with attaching edge on the near side'' or ``middle node with attaching edge on the far side''.}
\label{tab:logic_explanations}
\end{table}

\subsection{Interventions}
\begin{figure}[!t]
    \centering
    \includegraphics[width=0.6\textwidth]{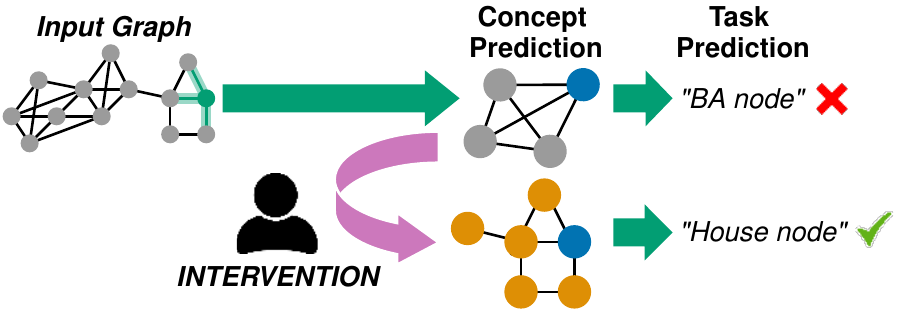}
    \includegraphics[width=0.38\textwidth]{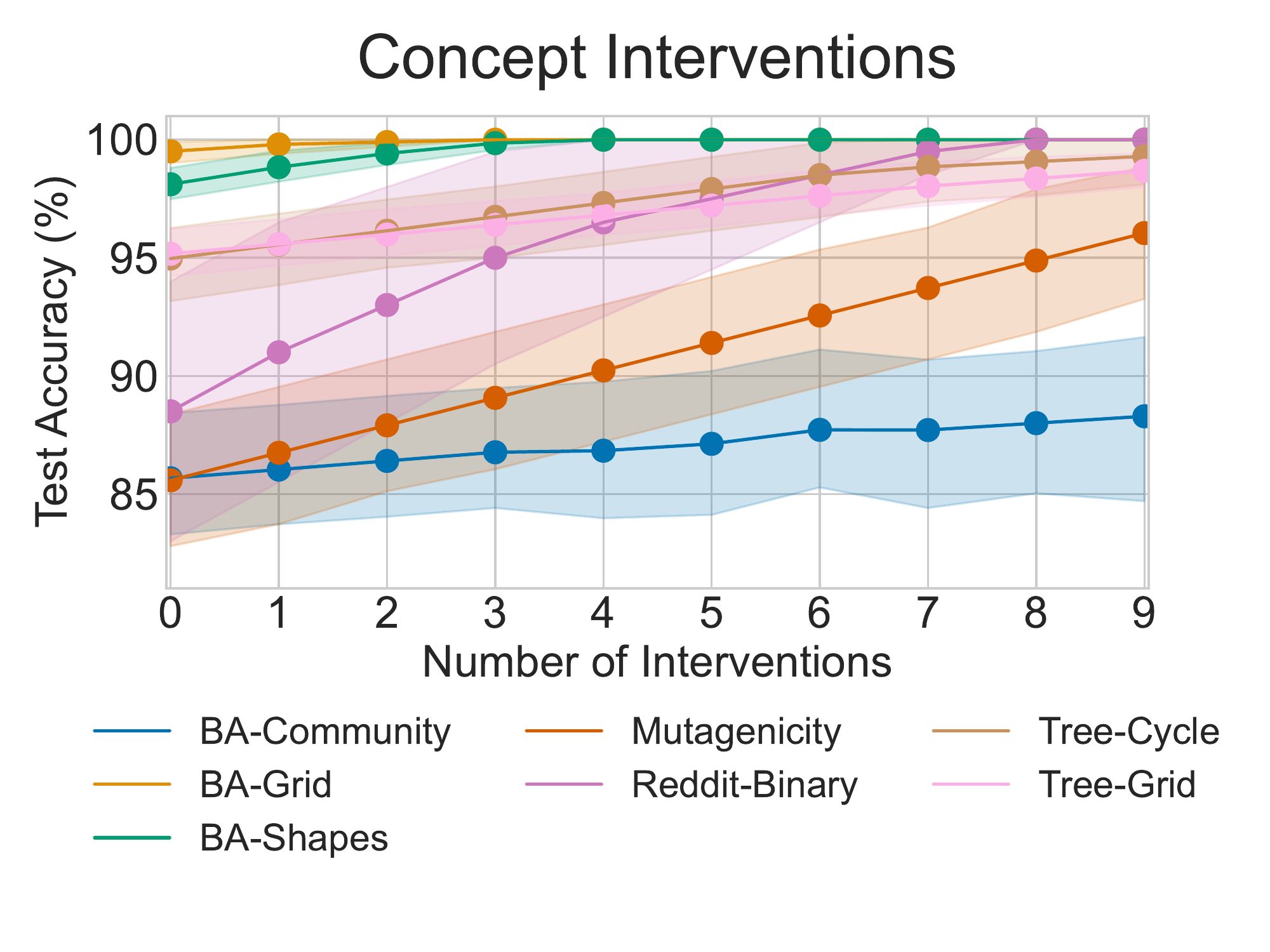}
    \caption{The Concept Encoder Module supports interventions at concept-level, allowing human experts to correct mispredicted concepts (left), increasing human trust in the model~\citep{shen2022trust}. This interaction  significantly improves task performance, achieving almost $100\%$ accuracy on synthetic datasets (right).}
    \label{fig:interventions}
\end{figure}

\paragraph{The Concept Encoder Module supports human interventions (Figure~\ref{fig:interventions})}
Supporting human interventions is one of the main benefits of more interpretable architectures that learn tasks as a function of concepts. In contrast to vanilla GNNs, CEM enables interventions at concept-level, which allows human experts to correct mispredicted concepts. Similarly to Concept Bottleneck Models~\citep{koh2020concept}, our results show that correcting concept assignments significantly improves the model test accuracy to over $98\%$ for the synthetic datasets, achieving $100\%$ test accuracy on BA-Grid and BA-Shapes. We also observe an increase in task accuracy in BA-Community, however, the increase is much more gradual. Most notably, in both real-world datasets CEMs allow GNNs to improve their task accuracy by up to $\sim + 10\%$ with less than $10$ interventions.

\section{Discussion} \label{sec:discussion}

\paragraph{Concept Graph Networks are accurate and self-explaining}
In summary, our results demonstrate that CEM makes GNNs explainable by design without impairing their task performance. Specifically, our approach extracts high-quality concept-based and logic explanations. Our experiments show that the extracted concepts are pure, meaningful and interpretable, while task-specific logic explanations are simple and accurate. We also demonstrate that CEM supports human interventions at concept level, which is one of the main advantages of explainable-by-design architectures.

\paragraph{Related work}
Graph Neural Network Explainer (GNNExplainer, ~\citep{ying2019gnnexplainer}) represents the first seminal work on GNN explainability. Specifically, GNNExplainer maximizes the mutual information between GNN predictions and the distribution of possible subgraphs for explanations. By focusing on individual predictions, the method is limited to explaining one instance at a time corresponding to a localized view of the data distribution. To get a full picture,~\citet{ying2019gnnexplainer} suggest to perform subgraph matching on a substantial number of instances. However, this is not scalable as subgraph matching is NP-Hard~\citep{ying2019gnnexplainer}. In an attempt to alleviate this issue, the Parameterised Graph Explainer (PGExplainer,~\citep{luo2020parameterized}) and the Probabilistic Graphical Model Explainer (PGM-Explainer,~\citep{vu2020pgm}) parametrize the process of generating explanations using deep neural networks to provide multi-instance explanations. However, all these methods remain fundamentally limited in their locality as they cannot explain a class of samples in its entirety. In contrast, GCExplainer ~\citep{magister2021gcexplainer} fills the gap of global explainability for GNNs using concept-based explanations. While the existing explainability techniques begin to address the lack of insight into the computations of GNNs, they are all post-hoc methods whose goal is to explain a trained GNN, not to make it more interpretable. The proposed method instead aims at filling this knowledge gap by making GNNs explainable by design. 

From a broader perspective, our work borrows ideas from supervised and unsupervised concept-based methods. These methods have been explored in various ways for other neural networks, such as convolutional neural networks~\citep{alvarez2018towards,ghorbani2019towards,chen2020concept,kazhdan2020now,koh2020concept} and recurrent architectures~\citep{kazhdan2020meme}. From supervised concept-based methods, our approach inherits the ability to perform effective human interventions at concept level extending Concept Bottleneck Models~\citep{koh2020concept} to graphs. As for unsupervised methods, our approach mainly draws from the Automatic Concept Extraction algorithm (ACE,~\citep{ghorbani2019interpretation}). The algorithm extracts visual concepts by performing k-Means clustering~\citep{forgy1965cluster} on image segments in the activation space of a convolutional network. The ACE approach is based on the observation that the learned activation space is similar to human perceptual judgement~\citep{zhang2018unreasonable}. This was the main motivation behind GCExplainer, as well as our approach. However, in contrast to ACE and GCExplainer, we embed the clustering step within the network architecture, making GNNs explainable by design. 

\paragraph{Strengths and limitations}
The main limitation of our work is the association of only one concept per sample. However, this also applies to GCExplainer, as well as to state-of-the-art unsupervised explainability methods for convolutional networks, such as ACE~\citep{ghorbani2019towards}. The second main limitation pertains the visualization technique inherited from GCExplainer. Visualizing a concept by simply exploring the $p$-hop neighborhood may include nodes which are not relevant for identifying a concept. More specifically, the concept visualization technique could be improved by performing largest common subgraph matching across the samples representing a concept. However, such an approach would be extremely expensive in terms of computations even for small graphs and it would not scale for large concepts. 
In terms of novelties, the proposed approach is the first of its kind in terms of making GNNs explainable by design. Secondly, the approach allows to find the optimal number of concepts dynamically. While the size of the embedding space must still be defined, this size is just an upper bound, alleviating the user from the burden of tuning this hyperparameter as in other explainability methods, such as GCExplainer or ACE. This dynamic adaptation often produces a high number of clusters/concepts. While a higher number of concepts may appear redundant and be more complex to reason about, the extracted concepts accurately describe the dataset, as indicated by the high concept completeness scores and rare concepts found. Moreover, logic-based formulas allow to filter through the concepts relevant for each class. We note, however, that as stronger interpretable models are deployed, there are risks of societal harm which we must be vigilant to avoid.


\paragraph{Conclusion}
In this work, we address the lack of human trust in GNNs caused by their opaque reasoning. To this aim, we propose the Concept Encoder Module which makes GNNs explainable by design. We demonstrate that the proposed method allows to discover and extract high-quality concept-based explanations. The proposed approach makes GNNs explainable by design without a reduction in performance, while also allowing for human intervention. Human intervention allows to alleviate dataset biases, further increasing trust in the model. The increased understanding of the model's working through the proposed approach fosters an increase in trust and may open up the possibility to use GNNs in more high-stake scenarios.

\bibliographystyle{unsrtnat}
\bibliography{references}

\newpage

\newpage

\appendix


\section{Experimental setup}

\paragraph{Model training and evaluation}
For each dataset, we train two types of models: a CGN, a model implementing the proposed approach, and a vanilla GNN. Both models have the same number of graph convolutional layers~\citep{Kipf2016GCNConv}, where we base the number of layers on those reported by ~\citet{magister2021gcexplainer} for GCExplainer. In the CGN, these layers are then followed by the CEM. In lieu of the CEM, the vanilla GNN has a linear layer and a log softmax activation function to perform the read out. We perform a grid search on the CGN for each dataset to determine the number of hidden units, as well as the learning rate. The number of hidden units tested for the CGNs for the synthetic datasets are 10, 20 and 30, while we search over 20, 30 and 40 hidden units for the real-world datasets, as they are more complex. The different learning rates tested are 0.1, 0.001 and 0.0001. In the more complex datasets Mutagenicity and Reddit-Binary, we also search over the concept encoding size to encourage fewer clusters. The concept encoding size is the number of hidden units in the final graph convolutional layer. We vary this parameter rather than using the same number of hidden units throughout, as we observed a high number of very small clusters formed for those datasets. This can be attributed to the real-world datasets having a less regular structure than the synthetic ones. We determine the best hyperparameter combination based on model convergence and test accuracy.  Table \ref{table:models} summarises the model architecture and hyperparameters defined for each dataset.

\begin{table}[h]
\resizebox{\textwidth}{!}{%
\begin{tabular}{|l|c|c|c|c|c|c|}
\hline
\textbf{Dataset} &
  \textbf{\begin{tabular}[c]{@{}c@{}}Number of \\ Graph Convolutions\end{tabular}} &
  \textbf{\begin{tabular}[c]{@{}c@{}}Number of \\ Hidden Units\end{tabular}} &
  \textbf{\begin{tabular}[c]{@{}c@{}}Concept \\ Encoding Size\end{tabular}} &
  \textbf{\begin{tabular}[c]{@{}c@{}}Learning \\ Rate\end{tabular}} &
  \textbf{\begin{tabular}[c]{@{}c@{}}Batch \\ Size\end{tabular}} &
  \textbf{\begin{tabular}[c]{@{}c@{}}Number of \\ Epochs\end{tabular}} \\ \hline
\textbf{BA-Shapes}     & 4 & 10 & 10 & 0.001  & N/A & 7000  \\ \hline
\textbf{BA-Grid}       & 4 & 10 & 10 & 0.001  & N/A & 3000  \\ \hline
\textbf{Tree-Grid}     & 7 & 20 & 20 & 0.0001 & N/A & 20000 \\ \hline
\textbf{Tree-Cycle}    & 3 & 10 & 10 & 0.001  & N/A & 7000  \\ \hline
\textbf{BA-Community}  & 6 & 20 & 20 & 0.0001 & N/A & 10000 \\ \hline
\textbf{Mutagenicity}  & 4 & 40 & 10 & 0.001  & 16  & 1000  \\ \hline
\textbf{Reddit-Binary} & 4 & 40 & 10 & 0.001  & 16  & 1000  \\ \hline
\end{tabular}%
}
\caption{A summary of the model architecture and training hyperparameters for each dataset. The number of hidden units and learning rate where determined via a grid search, while the number of graph convolutional layers is based on the models reported for GCExplainer~\citep{magister2021gcexplainer}. These parameters are kept the same between the Concept Graph Network and vanilla graph neural network to allow comparison.}
\label{table:models}
\end{table}

In regard to the initialization of GCExplainer, we carry over the recommended values for $k$ in k-Means proposed by ~\citet{magister2021gcexplainer}. The recommended values for $k$ in BA-Shapes, BA-Grid, Tree-Grid and Tree-Cycle are $k=10$, while $k=30$ for BA-Community, Mutagenicity and Reddit-Binary. ~\citet{magister2021gcexplainer} determine these experimentally by performing a grid search and comparing the concept completeness and purity scores for different initalizations of $k$. We also carry over the initialization of $p$ for visualizing the $p$-hop neighborhood, recommend by ~\citet{magister2021gcexplainer}. For BA-Shapes and BA-Community, $p=2$ to cover the  `house-motif`. For the remaining datasets, $p=4$ to visualise the encoded motifs.

For each dataset, we train and evaluate each of the models five times in order to be able to state confidence bounds. In order to allow reproducing the results, we fix the random seed for a run, using a different seed for each run. The randomly generated seeds for the five runs are: 42, 19, 76, 58 and 92.

\paragraph{Metrics}
Here, we will detail how we compute the five metrics for evaluating task accuracy, task completeness, concept interpretability and explanation performance. The first metric is classification accuracy, which we compute by dividing the number of correctly classified samples by the total number of samples. Secondly, we calculate concept completeness~\citep{yeh2020completeness}, which is the accuracy of a decision tree classifier trained to predict the output label from the concept encoding in the case of CGNs and cluster number in the case of GCExplainer. Thirdly, we compute concept purity. We follow the approach proposed by ~\citet{magister2021gcexplainer}, which is to calculate the graph edit distance between the top representation of a concept and the two closest concept representations to it. We compute the minimum purity score, because we want to compare the best concepts found by CEM and GCExplainer. As the graph edit distance is expensive to compute, only smaller, meaningful graphs are considered. Table \ref{table:purity_size} summarises and justifies the graph sizes considered. We aim for the maximum graph size to be twice the size of the concept, however, we cap this threshold at 13 if the concept is too big. This is reasonable, as in the synthetic datasets noisy concepts representing the base graph are identified.

\begin{table}[]
\resizebox{\textwidth}{!}{%
\begin{tabular}{|l|c|l|}
\hline
\textbf{Dataset} & \textbf{\begin{tabular}[c]{@{}c@{}}Graph Sizes considered\\ in Purity Score\end{tabular}} & \multicolumn{1}{c|}{\textbf{Comment}} \\ \hline
\textbf{BA-Shapes}     & 10 & Twice the size of the desired motif. \\ \hline
\textbf{BA-Grid}       & 13 & Capped due to computation cost.      \\ \hline
\textbf{Tree-Grid}     & 13 & Capped due to computation cost.      \\ \hline
\textbf{Tree-Cycle}    & 12 & Twice the size of the desired motif. \\ \hline
\textbf{BA-Community}  & 10 & Twice the size of the desired motif. \\ \hline
\textbf{Mutagenicity}  & 13 & Capped due to computation cost.      \\ \hline
\textbf{Reddit-Binary} & 13 & Capped due to computation cost.      \\ \hline
\end{tabular}%
}
\caption{A summary of the maximum graph size considered in the computation of the purity score.}
\label{table:purity_size}
\end{table}

Similar to the completeness score, we compute the accuracy of logic explanations by taking the accuracy of a decision tree classifier trained on mapping between concepts activated according to logic formula and the output label. Finally, we compute the complexity of logic explanations by measuring the length of the logic formula. The metrics reported are the mean of the five experiment runs for each dataset, including the $95\%$ confidence intervals computed using the Box-Cox transformation for non-normal distributions.

\paragraph{Implementation details}
The code for the experiments is implemented in Python 3, relying upon open-source libraries~\citep{pedregosa2011scikit,paszke2019pytorch,Fey/Lenssen/2019}. Implementation
can be found in the public repository \url{anonymous} using Apache v2.0 free and open source licence.
All the experiments have been run on the same machine: Intel\textsuperscript{\textregistered} Core\texttrademark\ i7-10750H 6-Core Processor at 2.60 GHz equipped with 16 GiB RAM and NVIDIA GeForce RTX 2060 GPU.

\section{Further concept visualisations}

In this section we present further concept visualisation results for the datasets. As an exhaustive number of concepts is identified for more complex datasets, such as BA-Community, Mutagenicity and Reddit-Binary, we only present a selection of the concepts discovered.

\begin{figure}[!t]
    \centering
    \includegraphics[width=\textwidth]{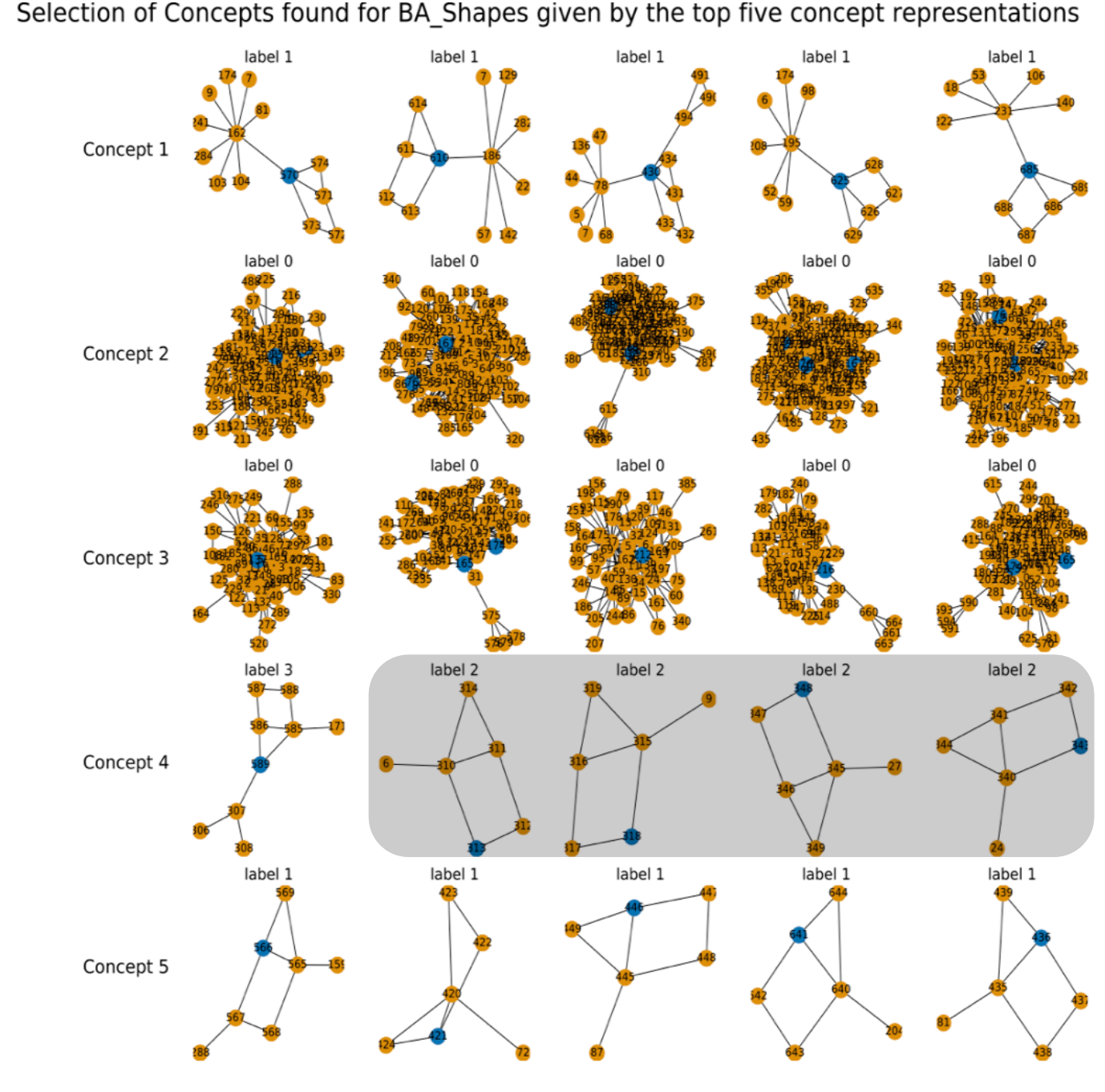}
    \caption{A selection of five concepts found for the BA-Shapes dataset, each represented by the top five concept representations. Blue nodes are the nodes clustered, while the orange nodes make up the $p$-hop neighborhood. The greyed out area shows that the cluster size for the concept is only 1. This means that only the first concept is representative. In contrast to the other concepts visualised, it becomes evident that this is a rare concept of a top of the house node being connected to the base graph via a random edge.}
    \label{fig:more_ba_shapes}
\end{figure}

\paragraph{BA-Shapes (Figure \ref{fig:more_ba_shapes}}
The selection of concepts presented illustrates that a variety of concepts is found. Concepts 1 and 5 show the middle node of the house structure. In contrast, the fourth concept visualised shows a rare concept. Focusing on the first concept representation of the concept shown, it is evident that the concept visualised is a house structure attached via two edges, of which one is a random edge. The size of the corresponding cluster is only one, which explains why the four other concept representations do not match. This draws the user's attention to the significance of the rare motif.

\begin{figure}[!t]
    \centering
    \includegraphics[width=\textwidth]{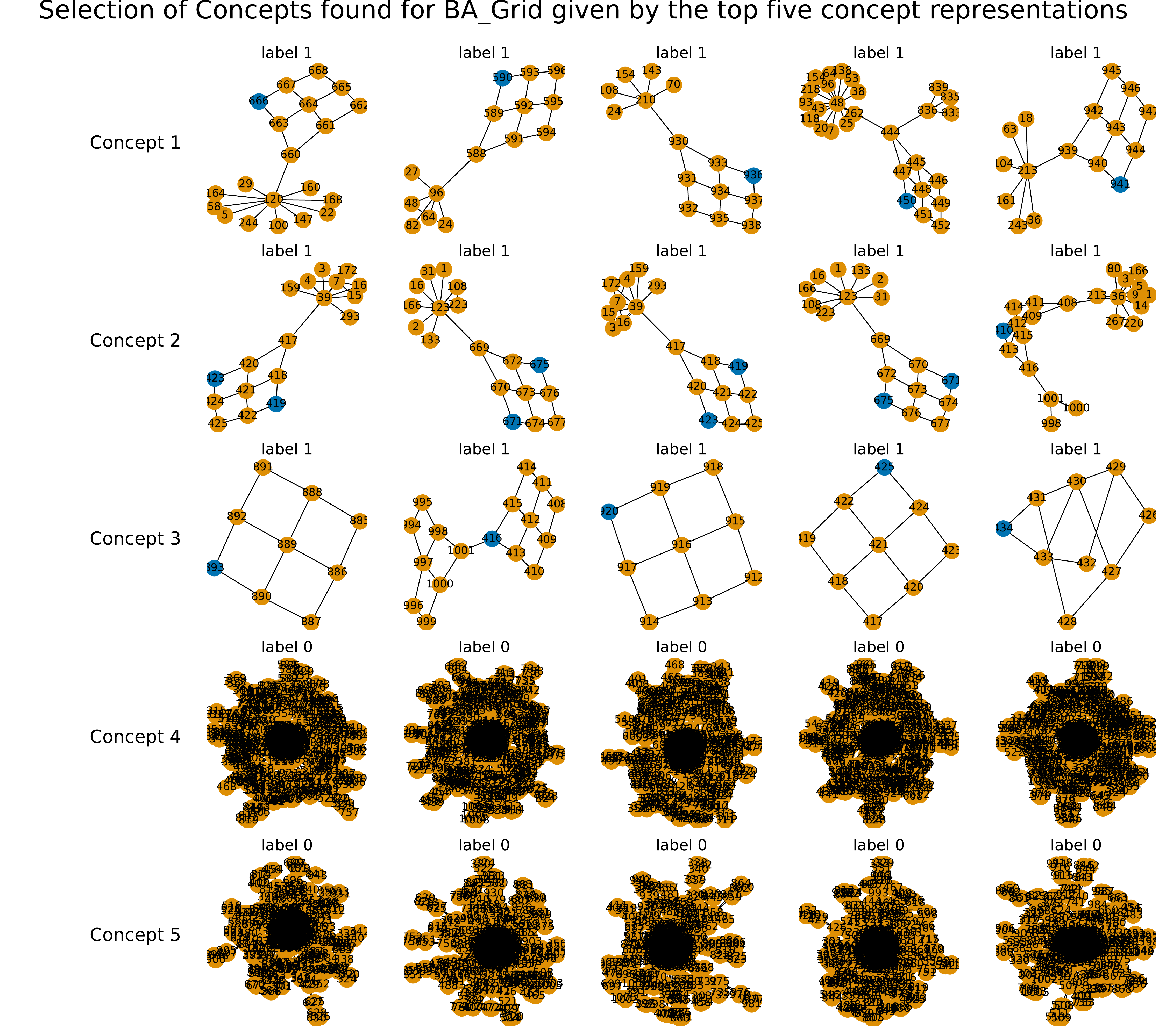}
    \caption{A selection of five concepts found for the BA-Grid dataset. Each concept is represented by the top five concept representations, the nodes closest to the center of a cluster. Blue nodes are the nodes clustered, while the orange nodes make up the $p$-hop neighborhood.}
    \label{fig:more_ba_grid}
\end{figure}

\paragraph{BA-Grid (Figure \ref{fig:more_ba_grid})}
We visualise a selection of the concepts found for BA-Grid in the same manner. In reference to concepts 1-3, it can be stated that the grid motif is successfully recovered. Moreover, the invariance of the graph is highlighted in concept 2, as we see two blue nodes per concept representation. Nodes on either side of the grid being grouped together shows that the GNN successfully classifies them as similar due to the neighbourhood having the same structure. In contrast, concepts 4 and 5 represent the BA base graph, which can be described as highly connected. 

\begin{figure}[!t]
    \centering
    \includegraphics[width=\textwidth]{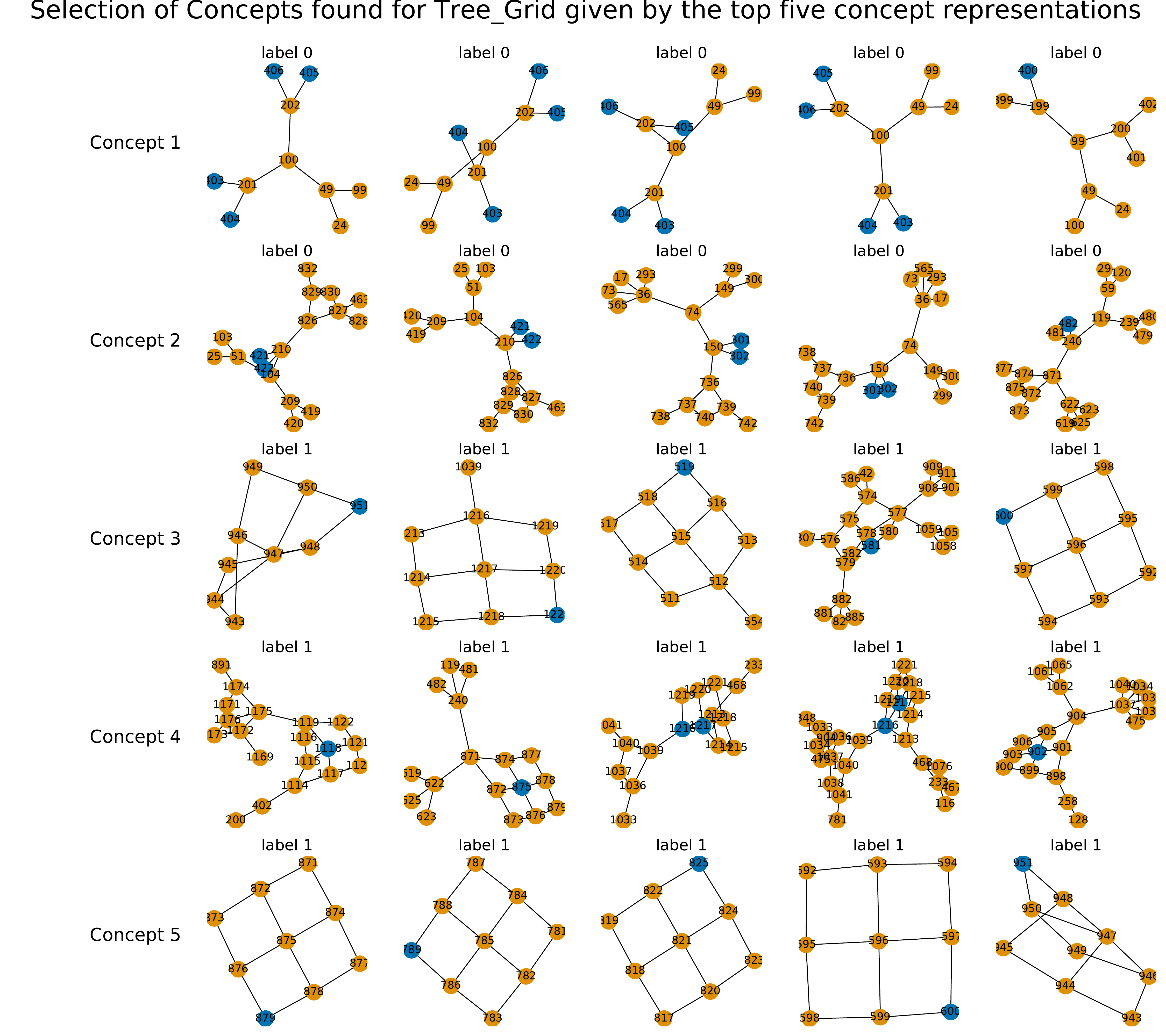}
    \caption{A selection of five concepts found for the Tree-Grid dataset, each represented by the top five concept representations. Blue nodes are the nodes clustered, while the orange nodes make up the $p$-hop neighborhood.}
    \label{fig:more_tree_grid}
\end{figure}

\paragraph{Tree-Grid (Figure \ref{fig:more_tree_grid})}
Similarly to BA-Grid, we successfully recover the grid concept, as shown in concepts 3-5. Moreover, concepts describing the more structured base graph are found. Concepts 1 and 2 clearly depict the binary tree structure, which forms the base graph, as well as parts of grid structures.

\begin{figure}[!t]
    \centering
    \includegraphics[width=\textwidth]{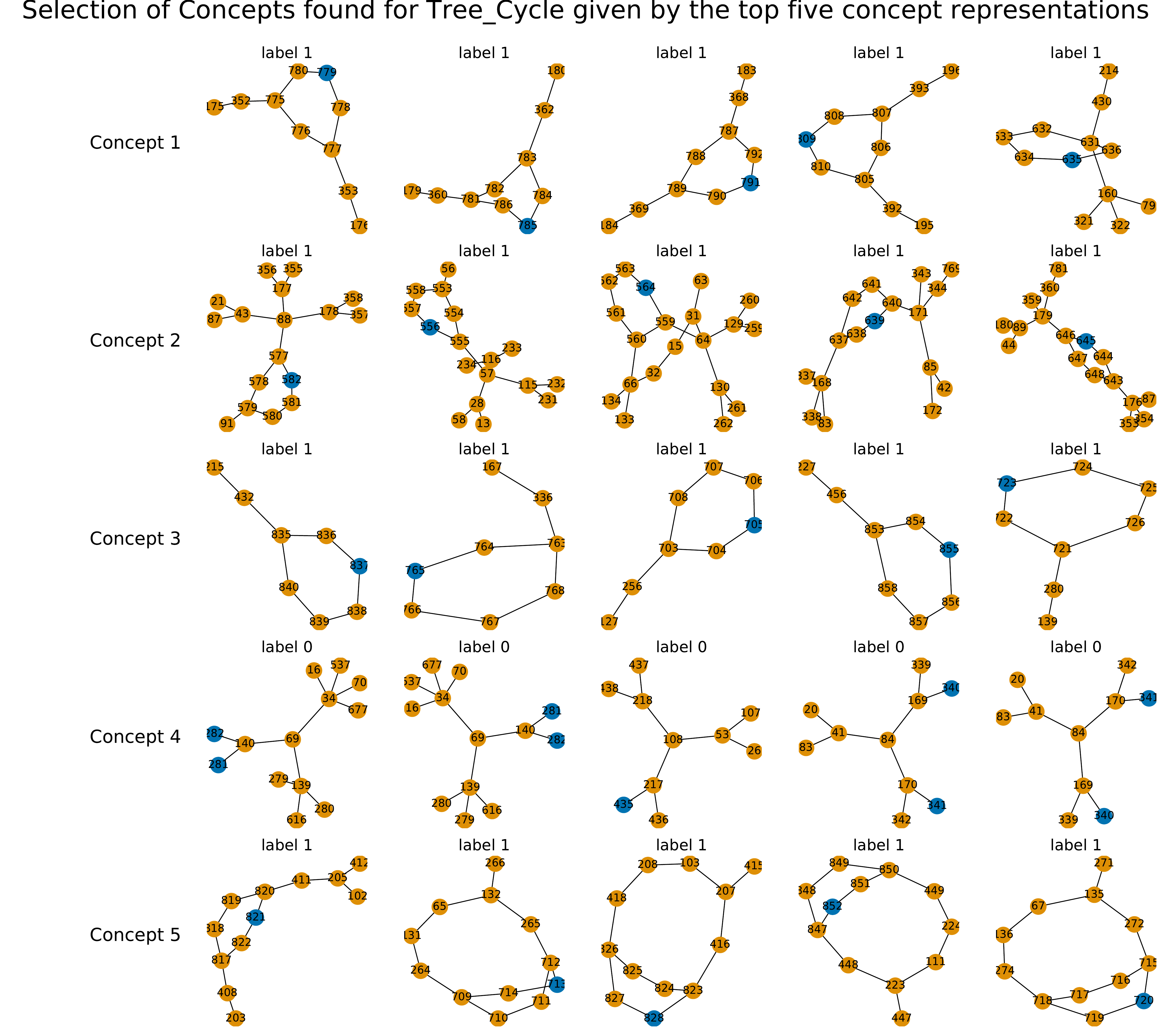}
    \caption{A selection of five concepts found for the Tree-Cycle dataset, each represented by the top five concept representations. Blue nodes are the nodes clustered, while the orange nodes make up the $p$-hop neighborhood.}
    \label{fig:more_tree_cycle}
\end{figure}

\paragraph{Tree-Cycle (Figure \ref{fig:more_tree_cycle})}
We obtain similar results for the Tree-Cycle dataset. The subset of concepts visualised in Figure \ref{fig:more_tree_cycle} shows that the cycle structure is successfully recovered in concepts 1-3. Concept 5 also represents the cycle structure, however, including random edges. Lastly, concept 4 encapsulates the binary-tree base graph of the dataset.

\begin{figure}[!t]
    \centering
    \includegraphics[width=\textwidth]{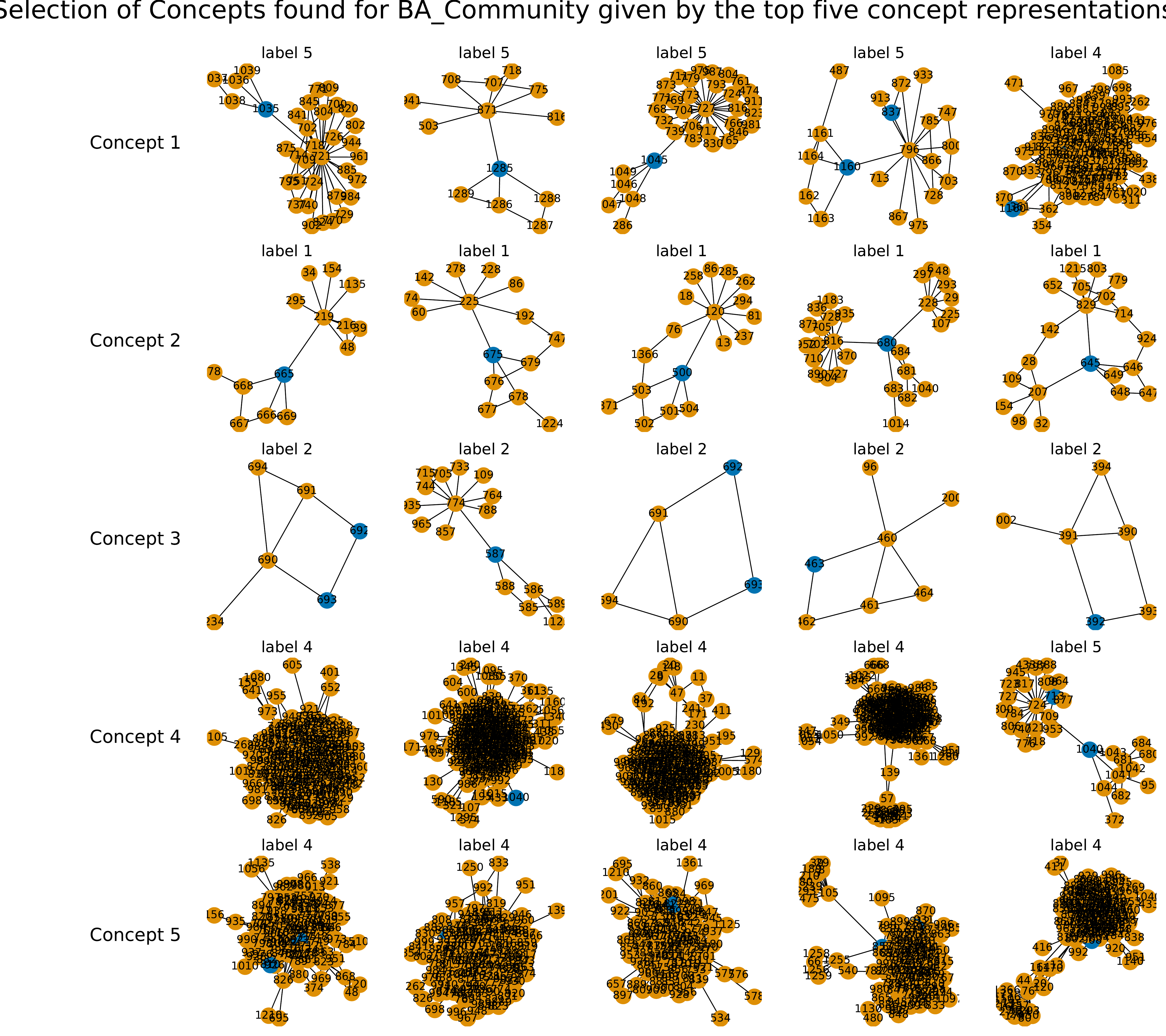}
    \caption{A selection of five concepts found for the BA-Community dataset, each represented by the top five concept representations. Blue nodes are the nodes clustered, while the orange nodes make up the $p$-hop neighborhood.}
    \label{fig:more_ba_community}
\end{figure}

\paragraph{BA-Community (Figure \ref{fig:more_ba_community})}
Figure \ref{fig:more_ba_community} shows concept representations of the concepts discovered for the BA-Community dataset. Concepts 1-3 successfully recover the house structure. However, concept 2 highlights the existence of random edges, creating rare structures. Finally, concepts 4 and 5 represent the highly connective nature of the BA base graph. Comparing concepts 1 and 2, it is evident that features play an important role and that feature importance is highlighted by CEM. Concept 1 and 2 both highlight the existence of the middle node attaching directly to the BA graph, however, reviewing the class labels, it is evident that house shapes belonging to different communities are clustered separated from one another. This indicates that the concept encoding differentiates between similar neighborhoods of the same structure based on feature importance.

\begin{figure}[!t]
    \centering
    \includegraphics[width=\textwidth]{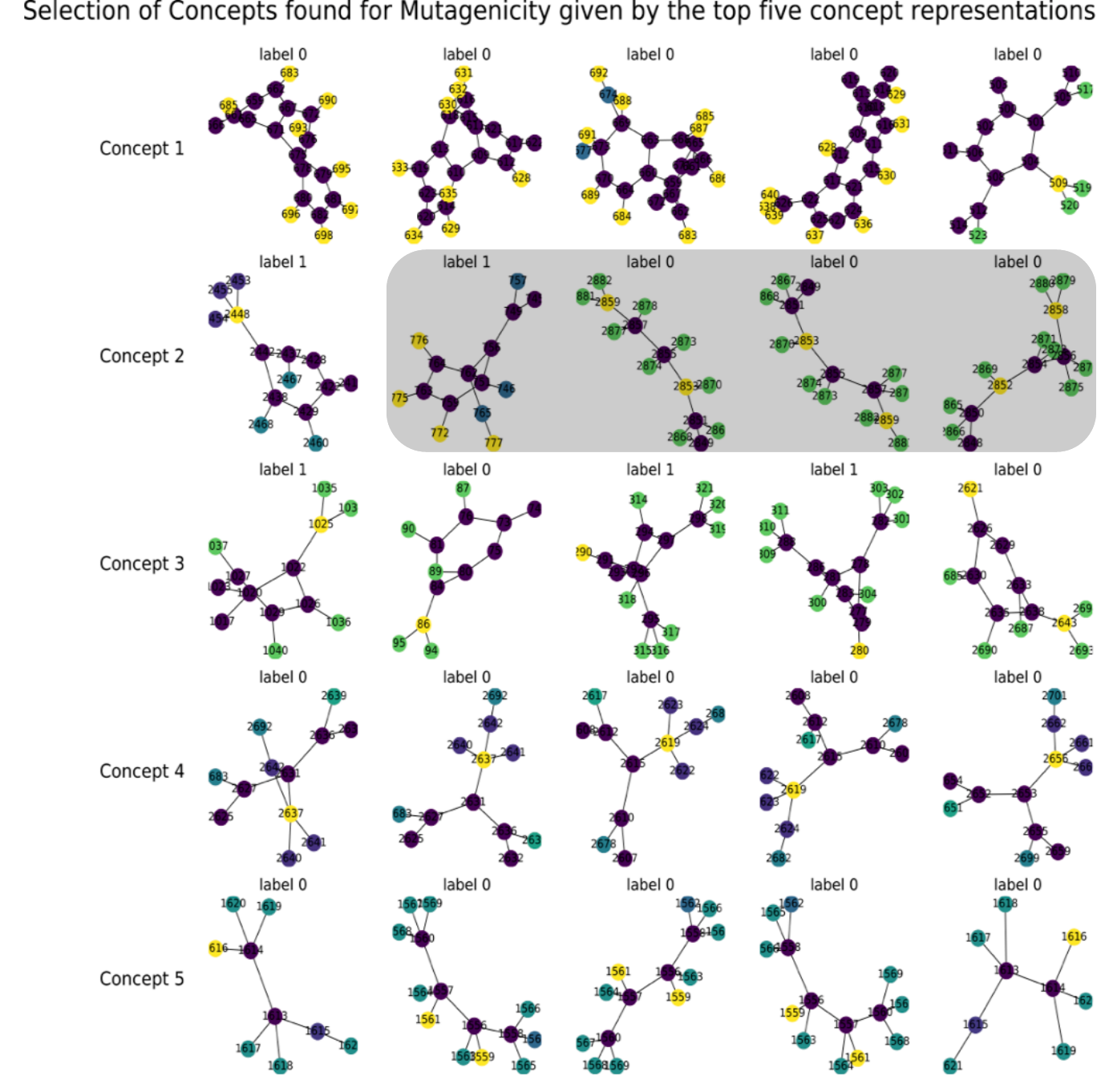}
    \caption{A selection of five concepts found for the Mutagenicity dataset, each represented by the top five concept representations. The nodes are colored by the node feature, which is the atom label. The greyed out area highlights a rare motif, as the cluster size is only one and there is a stark contrast to the other closest concept representations discovered.}
    \label{fig:more_mutagenicity}
\end{figure}

\paragraph{Mutagenicity (Figure \ref{fig:more_mutagenicity})}

Figure \ref{fig:more_mutagenicity} visualises a subset of concepts discovered for the Mutagenicity dataset. We can clearly identify the concept of the ring structure in concepts 1 and 3, while concepts 2, 4 and 5 exhibit a more varied structure, lacking the ring motif. Examining concept 3 more closely, it becomes evident that the ring structure does not uniquely identify a molecule as mutagenic. The concept depicts molecules which all have a ring structure, despite some being mutagenic and some not mutagenic, as indicated by the class label. Examining concept 2 in conjunction with the cluster size, it becomes evident that a rare concept is depicted. The size of the cluster is only one, which means that the greyed-out representations do not belong to this concept. Due to the lack of expert knowledge and ground truths, it is more tricky to identify the meaning of this rare concept, however, it can be stated that the second closest representation to the cluster centroid still exhibits the ring structure.

\begin{figure}[!t]
    \centering
    \includegraphics[width=\textwidth]{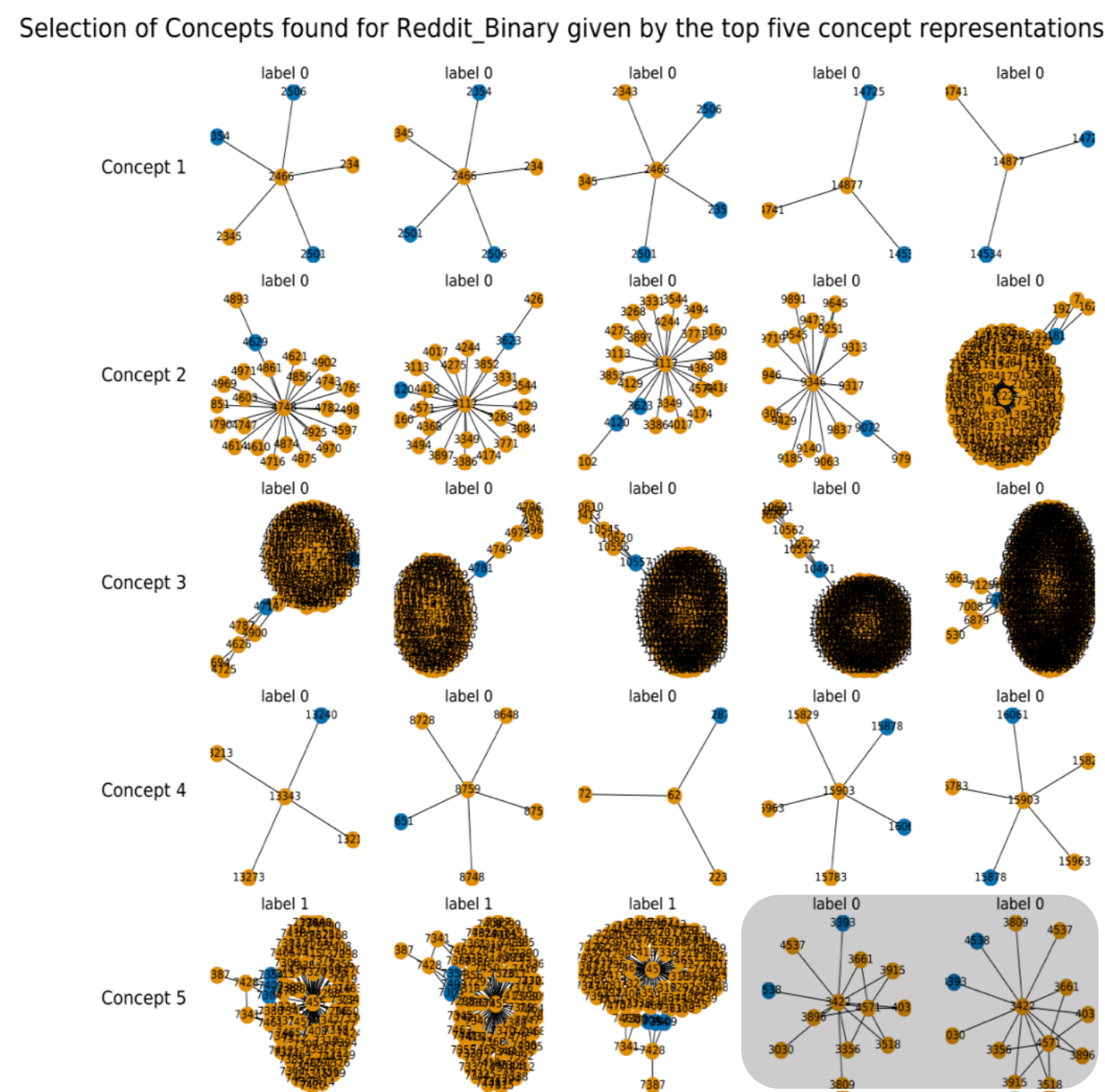}
    \caption{A selection of five concepts found for the Reddit-Binary dataset, each represented by the top five concept representations. Blue nodes are the nodes clustered, while the orange nodes make up the $p$-hop neighborhood. The greyed-out area highlights that the concept only comprises 3 examples and can be identified as a rare motif.}
    \label{fig:more_reddit}
\end{figure}

\paragraph{Reddit-Binary (Figure \ref{fig:more_reddit})}
Finally, Figure \ref{fig:more_reddit} visualizes a subset of the concepts found for Reddit-Binary. We can see that multiple concepts exhibiting the star-like structure are discovered, with the difference lying in the density of the connections. Examining concept 5 more closely, it can be stated that a rare concept is discovered, as there are only 3 examples of the concept. This may be attributed to the nodes not only being connected to the center node, marked by node 45, but also to other nodes, showing activity in other threads. The remaining two concepts visualised are the closest embeddings to the cluster centroid but are part of another concept.

\section{Comparison to GCExplainer and GNNExplainer}

We perform a qualitative comparison of the explanations produced using the CEM, GCExplainer~\citep{magister2021gcexplainer} and GNNExplainer~\citep{ying2019gnnexplainer}. We limit ourselves to a qualitative comparison against GCExplainer here, as we have already performed a quantitative comparison in the main section of the paper. We do not perform a quantitative evaluation against GNNExplainer, as the explanations are not concept-based and thus are evaluated in a different manner. GCExplainer and GNNExplainer are applied on the vanilla GNN. We focus on an evaluation of the BA-Shapes, BA-Grid and BA-Community datasets, as the ground truth motifs to be extracted are known for these dataasets.

\begin{figure}[!t]
    \centering
    \includegraphics[width=\textwidth]{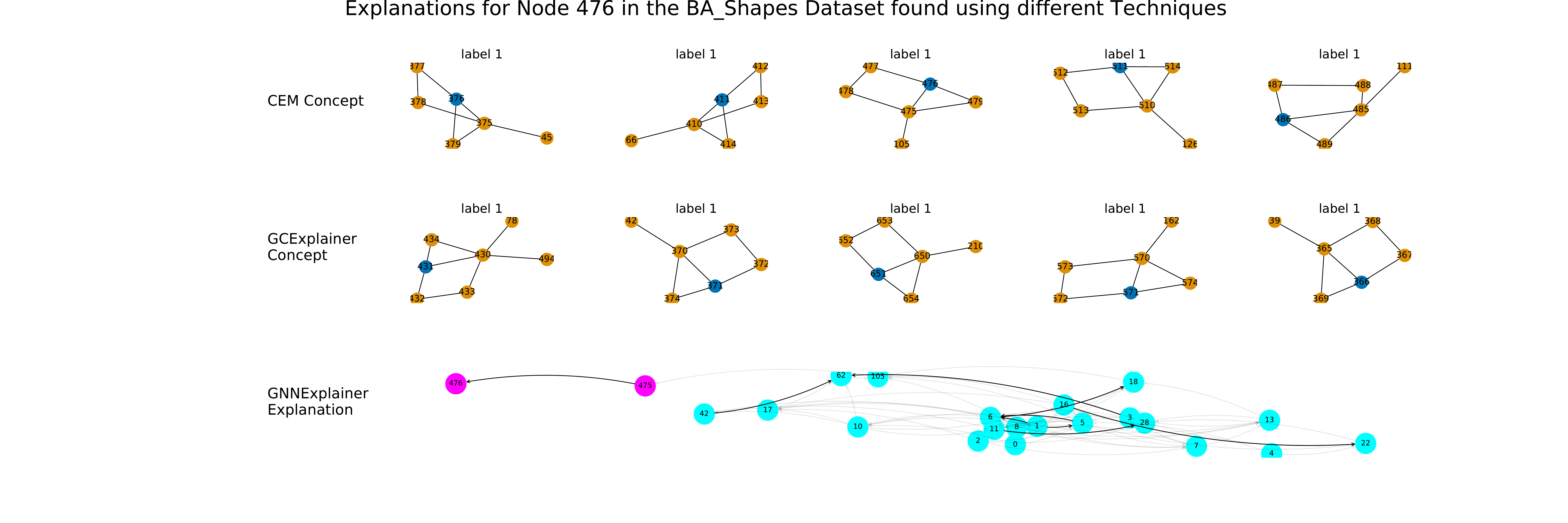}
    \caption{Concept-based explanations produced using the Concept Encoder Module and GCExplainer, as well as the explanation produced by GNNExplainer for a node in the BA-Shapes dataset. In the explanations, the blue nodes are the nodes clustered together, while the orange nodes are the $p$-hop neighborhood. GNNExplainer has its own coloring, where the purple nodes are node part of the middle of the house and the turquoise nodes are part of the BA base graph.}
    \label{fig:comp_ba_shapes}
\end{figure}

\paragraph{BA-Shapes (Figure \ref{fig:comp_ba_shapes})}
Figure \ref{fig:comp_ba_shapes} shows the explanations provides by CEM, GCExplainer and GNNExplainer for a node, which is part of the middle of a house. Both CEM and GCExplainer successfully identify the house structure. CEM performs slightly better than GCExplainer, as it does not include a concept with a random edge. In contrast, the explanation provided by GCExplainer does not visualise the house structure in full. Only the middle nodes of the house are visualised (purple), as well as a large part of the BA graph (turquoise). It can be argued that the explanations provided by CEM and GCExplainer are more intuitive, however, GNNExplainer highlights important edges.

\begin{figure}[!t]
    \centering
    \includegraphics[width=\textwidth]{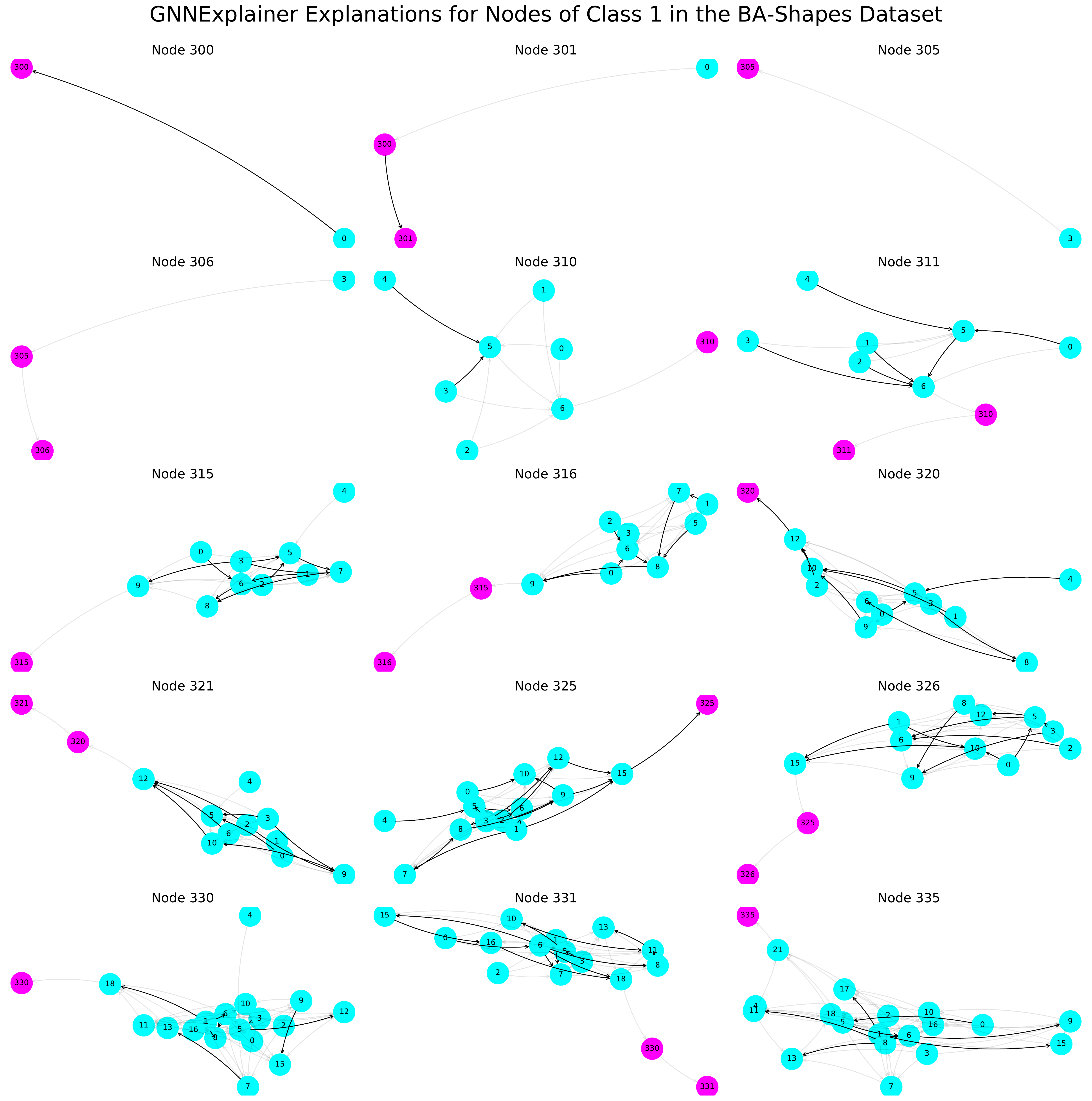}
    \caption{A selection of explanations produced using GNNExplainer for nodes of class 1. The purple nodes are nodes part of the middle of the house, while the turquoise ones are part of the BA base graph.}
    \label{bad_gnnexplainer}
\end{figure}

We struggle to reproduce the quality of explanations presented by ~\citet{ying2019gnnexplainer} for GNNExplainer. We first adapted the threshold to observe an effect on the explanations, however, this only impacted the visualisation of the important edges. We fix the threshold at 0.8 after this. We then examined the implementation of GNNExplainer used. To ensure that the quality of explanations is not the fault of the PyTorch Geometric~\citep{Fey/Lenssen/2019} implementation of GNNExplainer, we also used the implementation provided by the Deep Graph Library~\citep{wang2019deep}. After obtaining similar results, we exhaustively visualize the explanations for class 1. We present a selection in Figure \ref{bad_gnnexplainer}. In summary, we fail to produce the house motif using GNNExplainer, as the explanations provided mostly emphasize the importance of the BA base graph.

\begin{figure}[!t]
    \centering
    \includegraphics[width=\textwidth]{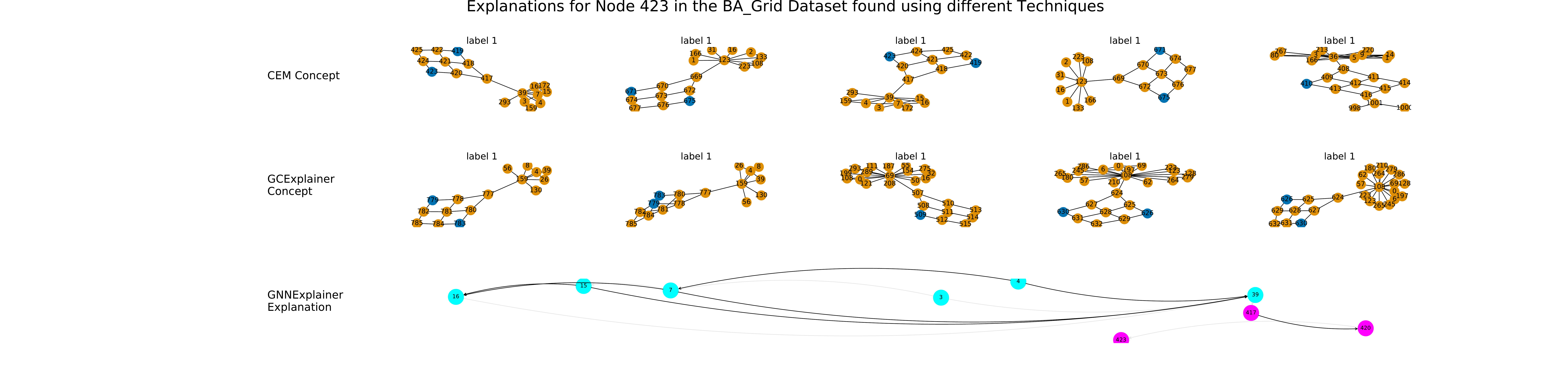}
    \caption{Concept-based explanations produced using the Concept Encoder Module and GCExplainer, as well as the explanation produced by GNNExplainer for a node in the BA-Grid dataset. In the explanations, the blue nodes are the nodes clustered together, while the orange nodes are the $p$-hop neighborhood. GNNExplainer has its own coloring, where the purple nodes are part of the grid and the turquoise nodes are part of the BA base graph.}
    \label{fig:comp_ba_grid}
\end{figure}

\paragraph{BA-Grid (Figure \ref{fig:comp_ba_grid})}
Figure \ref{fig:comp_ba_grid} visualizes explanations produced for a grid node in the BA-Grid dataset. Based on the concept representations, it can be argued that CEM and GCExplainer produce explanations of similar quality, as they both successfully extract the grid structure. This is also supported by the comparable concept completeness scores achieved (Figure \ref{fig:accuracy}), however, CEM produces more pure results (Figure \ref{fig:purity}). In respect to GNNExplainer, our previous claim of more intutive explanations is supported, as it is harder to reason about the explanation provided. It can be inferred that the immediate neighbours part of the grid structure (purple) are the most important for prediction, as well as part of the BA graph (turquoise).

\begin{figure}[!t]
    \centering
    \includegraphics[width=\textwidth]{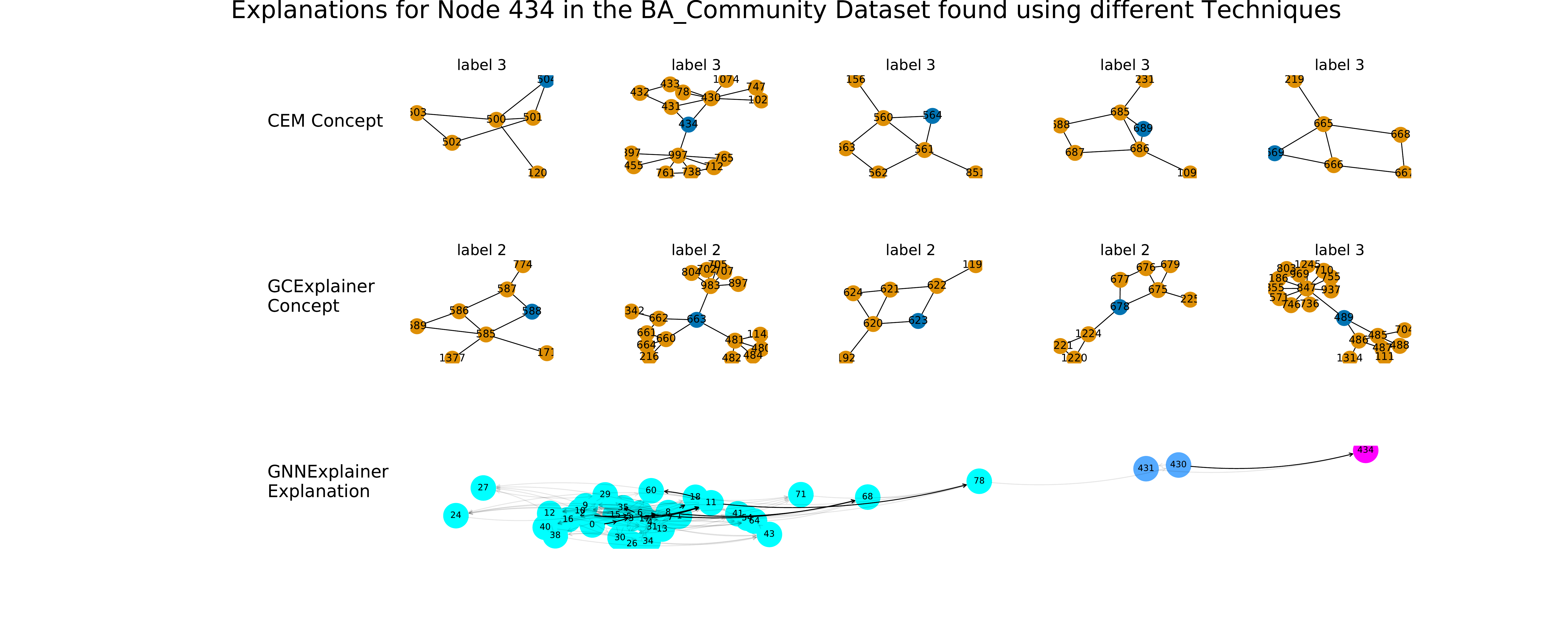}
    \caption{Concept-based explanations produced using the Concept Encoder Module and GCExplainer, as well as the explanation produced by GNNExplainer for a node in the BA-Community dataset. In the explanations, the blue nodes are the nodes clustered together, while the orange nodes are the $p$-hop neighborhood. GNNExplainer has its own coloring, where the purple node is the 'top' node in the house, the blue nodes are the 'middle' of the house and the turquoise nodes are part of the BA base graph.}
    \label{fig:comp_ba_comm}
\end{figure}

\paragraph{BA-Community (Figure \ref{fig:comp_ba_comm})}
Lastly, we compare the explanations for a node in the BA-Community dataset (Figure \ref{fig:comp_ba_comm}). Similar to our previous observations, both CEM and GCExplainer successfully identify the house structure. More importantly, they both identify the existence of random edges to explain the node. In contrast, the explanation provided by GNNExplainer is more elusive, highlighting mostly the BA base structure. In conclusion, it can be stated that the concept representations for CEM and GCExplainer are almost identical, which can be attributed to the same visualisation technique being used. However, we refer the read back to the quantative evaluation in the results section, which highlights the strengths of CEM over GCExplainer.

\section{Evaluation on further GNN architectures}

To demonstrate that CEM can be applied to any GNN architecture, we perform a quantative analysis of the performance of CEM using different convolutional embedding layers. We keep the experimental setup the same and only vary the type of graph convolutional layer. We trial the chebyshev spectral graph convolution (Cheb, ~\citep{defferrard2016convolutional}), graph isomorphism Network layer (GIN, ~\citep{xu2018GINConv}) and Graph SAmple and AGgregate layer (SAGE,~\citep{pal2020pinnersage}), as they are prominent graph convolutions, covering both spectral and spatial approaches. We showcase the results for the BA-Shapes and BA-Community dataset.

\subsection{Task Accuracy and Completeness}

\begin{figure}[!t]
    \centering
    \includegraphics[width=\textwidth]{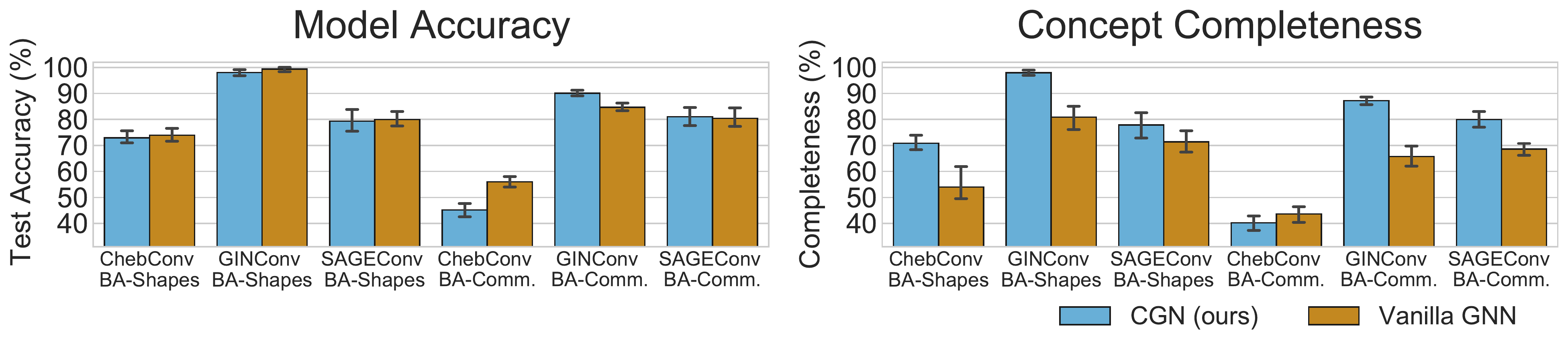}
    \caption{Model accuracy and concept completeness for the Concept-based Graph Network (CGN) and an equivalent vanilla GNN using different graph convolutional layers. For these results, and those that follow, we compute all metrics on test sets across five seeds and report their mean and $95\%$ confidence intervals.}
    \label{fig:accuracy_app}
\end{figure}

\paragraph{Concept Graph Networks are as accurate as vanilla GNNs (Figure \ref{fig:accuracy_app}, left)}
Our results confirm our previous observations. Across the three different layers, CEM allows to achieve better or comparable task accuracy w.r.t. equivalent GNN architectures. For example, the CEM-based model using GIN convolutions performs better on the BA-Community dataset. Only one CEM-based model performs worse than its respective vanilla GNN counterpart. That is the model trained on BA-Community. In general task accuracy is low, which can be attributed to the convolutional operation being ill-suited to the task.

\paragraph{The Concept Encoder Module discovers complete concepts (Figure \ref{fig:accuracy_app}, right)}
Our experiments confirm that CEM discovers a more complete set of concepts w.r.t. the concepts extracted by GCExplainer applied to the respective vanilla GNN. Most notably, CEM achieves significantly higher concept completeness scores on all runs, except for the Cheb-based model trained on BA-Community. However, the difference between the completeness score for concepts extracted using CEM and GCExplainer is not significant, despite a larger difference in model performance. This further confirms our conclusion that CEM discovers a more complete set of concepts than GCExplainer.

\subsection{Explanation performance}

\begin{figure}[!h]
    \includegraphics[width=\textwidth]{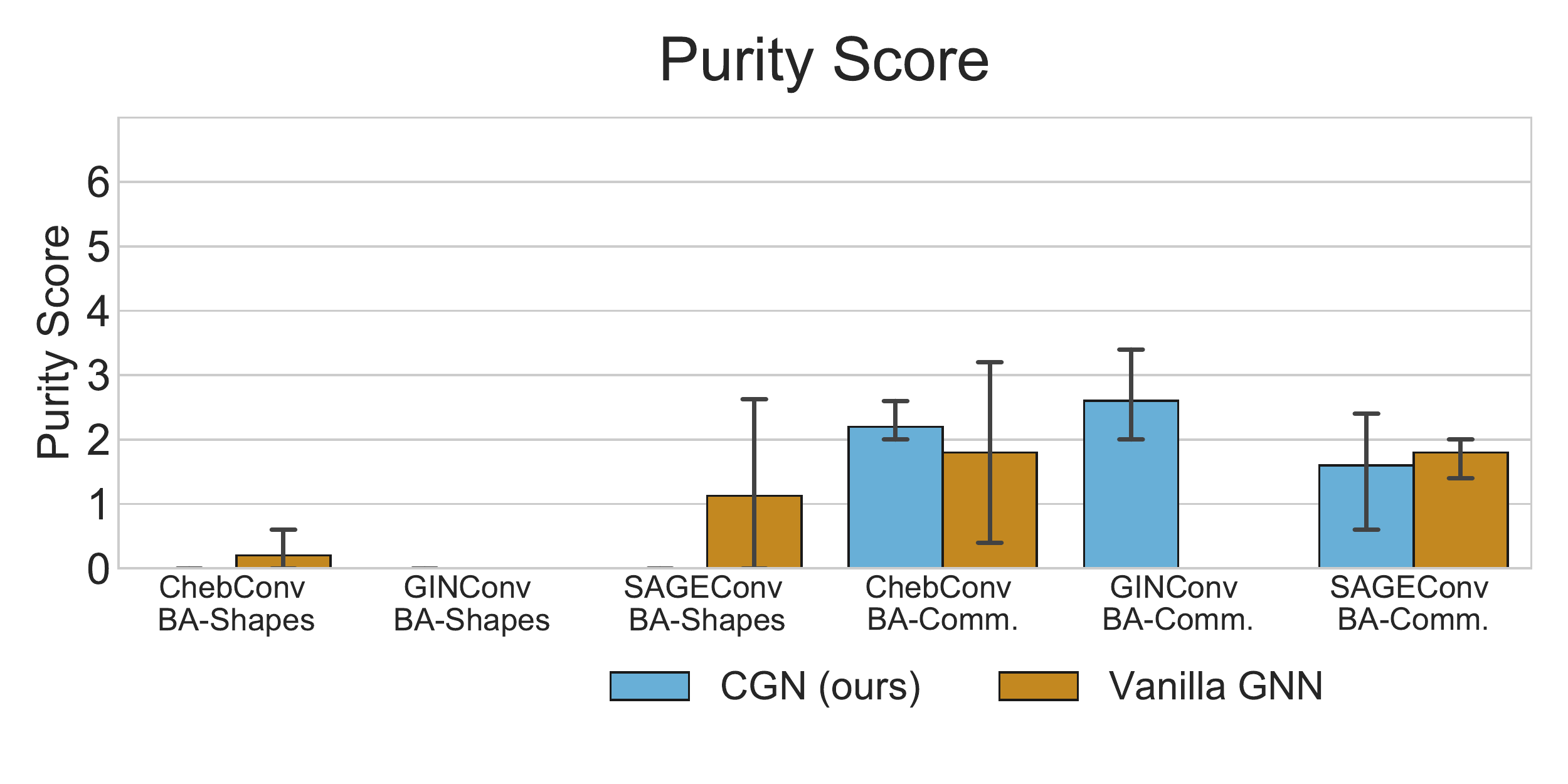}
    \captionof{figure}{Purity scores for the concept extracted by Concept Graph Module and GCExplainer. Notice how the optimal purity score is zero, as it measures the graph edit distance between concept instances~\citep{magister2021gcexplainer}.}
    \label{fig:purity_app}
\end{figure}

\paragraph{The Concept encoder Module identifies pure concepts (Figure \ref{fig:purity_app})}
We further demonstrate that CEM discovers high-quality concepts, which are coherent across samples. This is indicated by the low concept purity scores, where 0 is a perfect score. CEM outperforms GCExplainer applied on a vanilla GNN across all architectures on the BA-Shapes dataset. In contrast, it performs slightly worse on the BA-Community dataset, however, the variance of the purity score between the proposed approach and GCExplainer is not significant, with the exception for the model using the GIN convolution.

\begin{figure}[!h]
    \includegraphics[width=\textwidth]{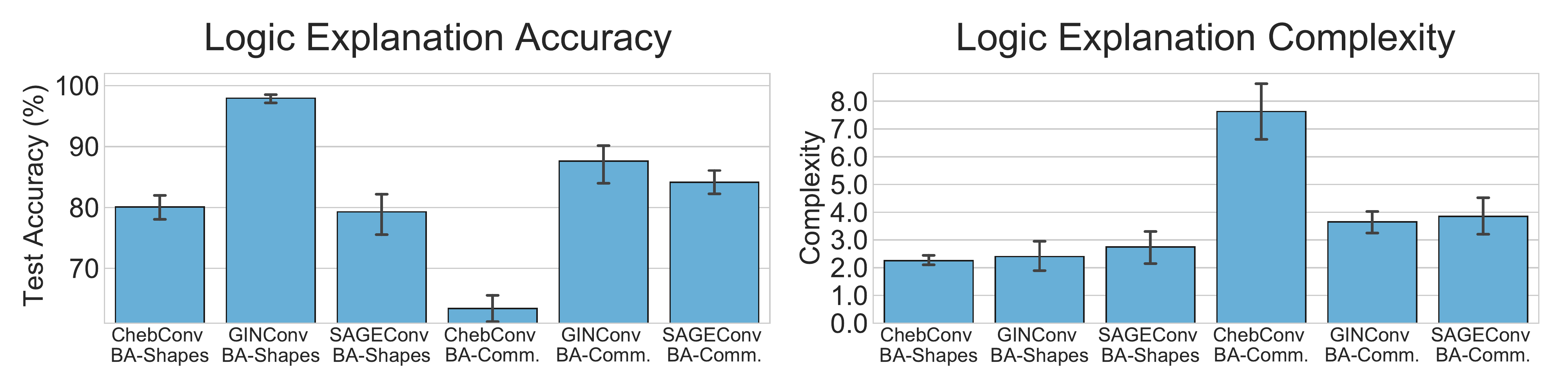}
    \captionof{figure}{Accuracy and complexity of logic explanations provided by the proposed Concept Graph Module. The accuracy is computed using logic formulas to classify samples based on their concept encoding. Explanation complexity measures the number of minterms in logic formulas.}
    \label{fig:lens_app}
\end{figure}

\paragraph{The Concept Encoder Module provides accurate logic explanations (Figure \ref{fig:lens_app}, left)}
Our results support our previous statement that CEM provides accurate logic explanations, as the logic explanation accuracy ranges at model accuracy. The low accuracy attained by the Cheb-based model trained on the BA-Community dataset can be explained by the poor model accuracy. The drop in the overall logic explanation accuracy in comparison to the results reported earlier can be lead back to model performance. Moreover, it should be highlighted that logic-based explanations are a benefit of CEM over GCExplainer.

\end{document}